# Autonomous Building Cyber-Physical Systems Using Decentralized Autonomous Organizations, Digital Twins, and Large Language Model

Reachsak Ly,[1] Alireza Shojaei, Ph.D.[2]

[1] Ph.D. Student, Myers-Lawson School of Construction, Virginia Polytechnic Institute and State University. ORCID: https://orcid.org/0000-0003-0332-1312. Email: reachsak@vt.edu
[2] Assistant Professor, Myers-Lawson School of Construction, Virginia Polytechnic Institute and State University, (corresponding author) ORCID: https://orcid.org/0000-0003-3970-0541. Email: shojaei@vt.edu

**Abstract**
Current autonomous building research primarily focuses on energy efficiency and automation. While traditional artificial intelligence has advanced autonomous building research, it often relies on predefined rules and struggles to adapt to complex, evolving building operations. Moreover, the centralized organizational structures of facilities management also hinder transparency in decision-making in the building operation management process, which limits the true building autonomy. Research on decentralized governing and adaptive building infrastructure, which could overcome these challenges, remains relatively unexplored. This paper addresses these limitations by introducing a novel Decentralized Autonomous Building Cyber-Physical System framework that integrates Decentralized Autonomous Organizations (DAOs), Large Language Models (LLMs), and digital twin to create a smart, self-managed, operational, and financially autonomous building infrastructure. This approach aims to enhance adaptability, enable decentralized decision-making, and achieve both operational and financial autonomy in building management. This study develops a full-stack decentralized application (Dapp) to facilitate decentralized governance of building infrastructure. An LLM-based artificial intelligence assistant is developed to provide intuitive human-building interaction for blockchain and building operation management-related tasks and enable autonomous building operation. The framework is validated through prototype implementation in a real-world building, with evaluations focusing on workability, cost, scalability, data security, privacy, and integrity. Six real-world scenarios were tested to evaluate the autonomous building system's workability, including building revenue and expense management, AI-assisted facility control, and autonomous adjustment of building systems. Results indicate that the prototype successfully executes these operations, confirming the framework's suitability for developing building infrastructure with decentralized governance and autonomous operation.

## 1. Introduction

Research on autonomous buildings has become a promising frontier in the field of smart and sustainable infrastructure. Autonomous buildings are characterized by their ability to operate independently through self-management, self-sufficiency, and intelligent operation. These buildings leverage a combination of advanced technologies to monitor, control, and automate the system's operation, reducing reliance on external resources and enhancing sustainability and efficiency. Existing research on autonomous buildings has primarily concentrated on achieving maximum energy efficiency, net-zero energy status, and ensuring energy self-sufficiency and off-grid performance. These objectives are undoubtedly crucial and have yielded significant advancements in the research on sustainable building operations. While energy autonomy [1] is crucial, it represents just one dimension of the overarching goal of achieving a comprehensive autonomous building infrastructure. To achieve operational autonomy within the building, researchers have further explored the integration of several advanced technologies, including building automation systems, artificial intelligence (AI), machine learning (ML), the Internet of Things (IoT), and digital twins, to enhance building intelligence and autonomy in facilities management by enhancing various building operation and system such as HVAC, lighting, energy management, etc. [2], [3], [4].

Machine learning techniques and AI have played a pivotal role in enabling these advancements. Their capacity to analyze large volumes of data, recognize trends, and make well-informed choices has contributed significantly to improving energy efficiency, enhancing occupant comfort, and optimizing building performance. However, the functionality of this conventional machine learning model may rely on predefined rules or specific training data, which, in some circumstances, may not adequately capture the dynamic and complex nature of building operations. Furthermore, they often struggle to adapt to evolving circumstances or incorporate contextual information effectively. The advent of large language models (LLMs) offers a promising avenue to overcome these limitations and unlock a new realm of possibilities for autonomous building operations. LLMs exhibit remarkable capabilities in natural language processing, which could enable seamless human-machine interactions and intelligent decision-making processes. Unlike conventional AI models, LLMs possess a deep understanding of contextual information and can engage in human-like conversations, allowing for more intuitive and adaptive control of building systems. Moreover, LLMs have the potential to facilitate complex decision-making processes by integrating multiple data streams, analyzing scenarios from diverse perspectives, and providing well-reasoned recommendations and adjustments to the building system. Its human-like reasoning capabilities can open up new avenues for human-building interaction research and enhancement of the building infrastructure autonomy.

Furthermore, the rise of Web3 technologies with the emergence of distributed ledger technologies (DLT) [5] , such as blockchain technology [6] and decentralized autonomous organization (DAO), [7] has introduced new paradigms of decentralized governance and decentralized finance (DeFi) in the built environment, which hold significant potential for enhancing the overall autonomy of building systems [8]. Blockchain (BC) is a digital public ledger that has all its data documented and stored in a transparent and tamper-resistant manner in a decentralized network. In addition, DAO is a digital and community-driven entity running on a blockchain network that functions transparently and autonomously with democratic and collective decision-making capabilities among its members [9]. In addition to being energy self-sufficient, multiple research studies suggest that next-generation building infrastructure also needs to become both operationally and financially autonomous with the capability of decentralized self-governance and self-ownership to become fully autonomous. For instance, Chang and Joha [10] also introduced the concept of 'common infrastructure,' a social practice that emphasizes collaborative production, sharing, and maintenance of resources within building infrastructure. They proposed the idea of a civic, self-owned, and autonomous infrastructure with blockchain technology to promote inclusivity and reduce the need for direct ownership and control over the building infrastructure. Wang et al. [8] also proposed the engineered ownership concept by proposing a blockchain-based system with automation capabilities with distributed rights and power shared between autonomous agents in building infrastructure. Similarly, Hunhevicz et al. [11] have also contributed to the research on the self-governing of physical space by proposing the concept and prototypes of decentralized autonomous space (DAS) using DAO.

However, previous works on self-governing infrastructure have yet to incorporate advanced technologies such as digital twin and AI into their proposed decentralized systems. Specifically, the research on the integration of DLT with digital twins and LLMs in the context of physical infrastructure remains unexplored. By integrating the context-aware capabilities and human-like intelligence of LLM with the decentralized governance and self-ownership framework offered by DLT, researchers can potentially explore new frontiers in building autonomy research and pave the way for the creation of a decentralized and self-governed, intelligent, and sustainable, autonomous building infrastructure.

To address this knowledge gap, this paper introduces a novel decentralized autonomous building cyber-physical system (DAB-CPS) framework that integrates DAOs, LLMs, and a digital twin to create a smart and autonomous building infrastructure. The specific research objectives of this study are the following:
1) To design and develop a decentralized autonomous building cyber-physical system framework for collective and decentralized governance of building infrastructure with AI-enhanced building operation.
2) To explore the potential of financially autonomous physical space by designing a revenue generation model for building spaces through a DAO-based decentralized platform.
3) To assess the scope and capacity of AI virtual assistant and agent in enhancing the human building interaction and building operation automation by developing an LLM-based virtual assistant for building system control.
4) To investigate the capability of generative AI in the web3-based decentralized governance by integrating the LLM-based virtual assistant in the DAO-based decentralized platform to assist DAO members and users in executing blockchain transactions and participating in the DAO's governance processes.
5) To develop a full-stack decentralized application (Dapp) to facilitate the governance of the autonomous building infrastructure by the DAO members and enhance user interactions with the physical space.

The contributions of this study to the existing body of knowledge are as follows: (1) Presenting a novel AI and blockchain-integrated decentralized governance framework model within the context of autonomous building, paving the way for a new paradigm in smart building management and operations (2) Demonstrating the feasibility and benefits of large language models in enhancing human-building interaction, intelligent decision-making, and automation of building operations. (3) Developing a full-stack, open-source Dapp that serves as a template for other blockchain and AI-related applications in decentralized governance for physical infrastructure and autonomous building. (4) Demonstrating the practical prototype and evaluation of a decentralized framework for autonomous building in an actual physical building environment (5) Providing perspectives on the challenges and opportunities related to incorporating generative AI models, such as LLMs, into decentralized governance frameworks that contribute to the understanding of the interplay between AI and Web3 technologies in the built environment.

The rest of this paper is organized as follows: Section 2 presents a comprehensive background on the relevant concepts and studies related to the research. This section first covers the fundamentals of blockchain technology and DAO before reviewing the literature on DAO and artificial intelligence integration. This section also reviews the related studies on the application of LLMs in the construction industry, as well as discusses DAO application in the built environment. The research method of the study is presented in section 3, while section 4 describes the proposed decentralized autonomous building cyber-physical system (DAB-CPS) framework. Section 5 provides the implementation and validation of the developed system through a real case study. Then, a discussion on the findings, implications, and limitations of the research, as well as the evaluation of system usability and performance, are made in section 6. Finally, the conclusion is presented in Section 7.

## 2. Literature review
### 2.1. Overview of Decentralized Autonomous Organization

A decentralized autonomous organization (DAO) is a novel organizational structure built on blockchain technology, characterized by its community-driven governance, transparent operations, and autonomous execution of decisions. DAOs function as digital entities operating on a decentralized blockchain network, enabling collective decision-making processes among their members while adhering to predefined rules encoded in smart contracts [7]. Smart contracts are self-executing programs implemented on the blockchain that activate automatically when certain predefined conditions are satisfied [12]. This innovative approach to organizational governance fundamentally diverges from traditional centralized structures, introducing a paradigm shift towards decentralized and autonomous operations.

One of the earliest and most well-known implementations of DAO was "The DAO," which was launched on the Ethereum network in 2016. Although it was eventually hacked and shut down, it sparked significant interest and paved the way for further exploration and development of DAO applications. As of June 2024, DeepDao analytics [13] reports a total of 2,437 active DAOs with around 3.2 million active weekly users, which collectively manage approximately $30 billion worth of blockchain assets. These DAOs engage 3.2 million active users weekly. Among these organizations, 213 DAOs hold treasuries exceeding $1 million, 113 DAOs have assets over $10 million, and 36 DAOs manage funds surpassing $100 million.

DAOs are built upon three fundamental principles: decentralization, autonomy, and automation [7]. In contrast to hierarchical management systems, DAOs employ a decentralized network architecture, which eliminates the centralized governing body [14]. In addition, governance within a DAO occurs through a democratic process, where community members collectively participate and vote on proposals [15]. Approved proposals are then autonomously executed by the smart contracts, ensuring consistent and transparent implementation of the collective decisions. Furthermore, the immutable and transparent characteristics of blockchain technologies allow DAOs to automate organizational processes and transactions through predefined rules encoded in smart contracts, fostering trust and accountability [16].

The governance process within a DAO typically involves the submission of proposals by community members, followed by a voting process where token holders can cast their votes. Once the voting period concludes, the proposal is either accepted or rejected based on the predetermined voting rules and thresholds. Accepted proposals are then automatically executed by the smart contracts, ensuring transparency and adherence to the agreed-upon rules.

In recent years, DAOs have attracted growing research interest within industry and academia across various disciplines, including decentralized finance (DeFi) [17], healthcare [18], and education [19]. However, there have been DAO implementations for decentralized governance in the built environment, especially in the building infrastructure domain.

### 2.2. Integration of DAOs and artificial intelligence

The idea of integrating AI and DAOs for decentralized governance was proposed by McConaghy, [20] where he explored the potential of AI DAOs and their societal implications. The synergy of artificial intelligence (AI) and DAOs holds immense potential to revolutionize decision-making processes, solve governance-related challenges, and unlock new opportunities across various domains, including the built environment. Traditional DAOs operate based on predefined smart contract codes and voting mechanisms. However, by leveraging the ability of adaptive machine learning and feedback loops, AI-powered DAOs can learn, make decisions, and automate administrative tasks that are typically performed by humans [21].

Experts and researchers have explored various scenarios of AI and DAO integration to illustrate their potential symbiotic relationship and practical applications (Fig. 1). Revoredo [22] presents different AI DAO use cases such as managing blockchain treasuries and assets, leveraging swarm intelligence for meta-governance, smart contracts automation, assisting DAO governance, and AI-generated content or services for DAO. One notable example is Singularity DAO, which leverages AI to facilitate asset management and investment decisions based on user behavior analysis [23]. In the realm of smart transportation, scholars suggest the use of intelligent agent networks to coordinate and optimize autonomous mobility systems through collaborative decision-making [24]. Furthermore, within the domain of smart cities, researchers have also proposed the integration of DAOs and parallel systems to facilitate the deployment of AI applications [25], by fusing the physical and virtual environments with spatial symbiotic intelligence, an integrated and collaborative DAO-AI decision-making system could be created to address the diverse needs of city residents more effectively [26]. Moreover, generative AI has been successfully integrated with DAOs to create self-sustaining digital organizations whose revenue is generated by AI-generated content. For instance, the works by Yadlapalli et al. [27] harnessed the potential of combining AI and DAOs by creating a digital, self-governing organization that sustains itself through the revenue generated from selling digital art automatically created by a Generative Adversarial Network (GAN). Similarly, Guo et al. [28] integrated artificial intelligence and DAOs to build a human-machine collaborative painting system for artistic content creation, further demonstrating the synergies between these two technologies. However, despite these advancements, there are a few research gaps that need to be addressed. First, most previous research has primarily focused on using AI to power DAO-related products and services through content generation or service provision. None have effectively utilized AI as a tool for assisting the DAO governance process. The few studies that have explored this area have only proposed conceptual frameworks but have yet to conduct empirical implementation. Furthermore, the integration of advanced AI models like LLMs with DAOs has yet to be explored. This presents a unique opportunity to investigate how LLMs can enhance DAO governance, decision-making processes, and overall efficiency. Addressing these gaps could pave the way for more sophisticated, intelligent, and autonomous decentralized systems.

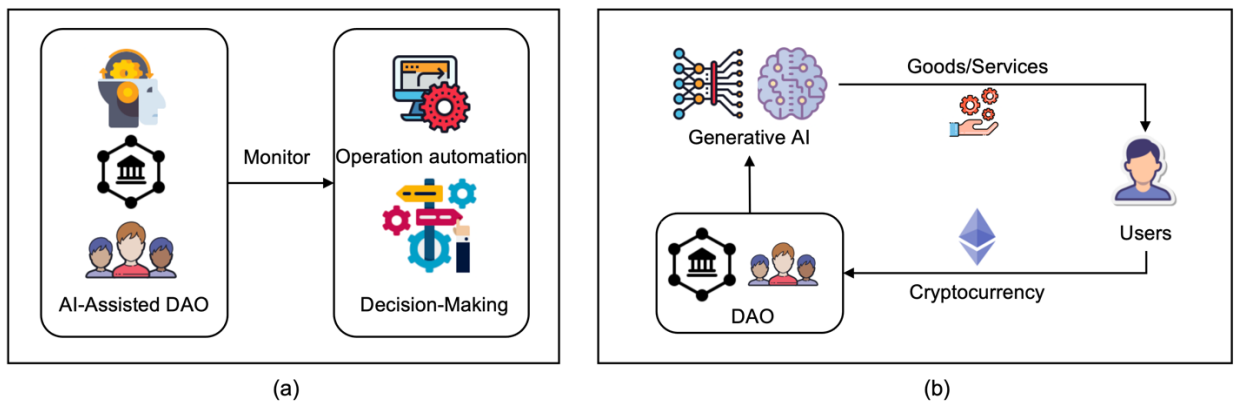

Fig 1. Integration of DAO and AI

### 2.3. Application of Large language model in construction industry

Large language models have emerged as a major advancement in the field of natural language processing (NLP) by revolutionizing the way humans interact with computer systems. These models are trained on large amounts of data, which allows them to comprehend and generate human-like language with remarkable fluency and coherence. Among the most prominent LLMs, OpenAI's ChatGPT has garnered significant attention due to its impeccable performance in information extraction and text summarization. These models leverage transformer-based architectures that are capable of learning patterns of natural language, which allow the generation of contextually relevant and coherent text by providing prompts that combine task instructions with selected contextual examples. A primary benefit of GPT models is their ability to generate language that closely resembles text written by humans. This has opened up a wide range of applications, such as chatbots and natural language communication, where these models can provide accurate and human-like responses to open-ended queries.

The capability and application of LLM have gained significant attention from researchers in the construction domain in the past few years. As a result, several studies have been conducted to explore the application of LLM in various stages of the construction project lifecycle, including project planning, construction, operations, and maintenance. For instance, in the project planning phase, a study by Prieto et al. [29] examines the use of LLM in construction schedule generation based on the project scopes and requirements. The findings show that the LLM developed a structured schedule that adopts a logical framework to meet the specified scope requirements. Within the construction phase, researchers have also explored the use of LLM in the construction robotics domain. You et al. [30] introduced RoboGPT, a novel system that utilizes the reasoning abilities of ChatGPT for automating sequence planning in robot-assisted assembly for construction tasks. Experimental evaluations and case studies demonstrated that RoboGPT-enabled robots could manage complicated construction operations and adapt to real-time changes. Additionally, LLM has also been used in the construction safety domain. Chen et al. [31] proposed a construction safety query system, which combines image captioning with a visual question-answering capability on head-mounted AR devices by leveraging vision language models such as ChatGPT 4. The system allows construction workers to query safety-related information by capturing images and asking questions through natural language interaction, thereby improving safety inspection, compliance checking, and progress monitoring in construction sites. In another study, Uddin et al. [32] examined the impact of integrating ChatGPT into the construction education curriculum. The investigation involved measuring students' hazard recognition abilities before and after introducing ChatGPT as an educational tool. The results showed significant improvements in hazard recognition, suggesting the potential benefits of integrating LLMs into construction safety education and training. Furthermore, LLM has also been used by researchers to facilitate efficient information search. For instance, Zheng and Fischer [33] introduced an AI-powered virtual assistant system that integrates ChatGPT for supporting natural language-based building information models (BIMs) search. This system allows users to query BIM databases using natural language, extract relevant information, and receive responses along with 3D visualizations.

While previous research has investigated the integration of GPT models in various domains of construction, the application of LLMs in the context of smart building infrastructure remains relatively unexplored. The advanced reasoning capabilities and natural language processing power of LLMs hold significant potential for enhancing automation and intelligent decision-making for smart building operations, as well as facilitating intuitive interaction between occupants and facility management operations. Furthermore, it is important to note that the existing research on the application of LLMs in the construction domain has primarily utilized commercialized GPT models, such as OpenAI's ChatGPT. Saka et al. [34] highlights some key concerns regarding data privacy, cost scalability, and confidentiality issues related to the usage of these models. The use of these cloud-based GPT models requires sensitive data to be transmitted to external servers for processing, potentially exposing it to unauthorized access or data breaches. Another significant challenge lies in the cost and scalability aspects of these commercial LLM services. As an example, users must pay a monthly subscription fee of $20 to access the ChatGPT web interface, which also comes with certain usage limitations. Similarly, the developed applications that use the GPT models API also face recurring charges based on their level of usage. These cost considerations may hinder the broader adoption and scalability of such applications within the construction industry. Therefore, there is a need to explore alternative solutions, such as utilizing local and open-source LLMs that can improve inference speed while addressing data privacy concerns, as the data remains localized within the device or system.

### 2.4. Decentralized autonomous organization and self-governing entity.

In recent years, various studies and prototypes have highlighted the potential of utilizing DAO in the development of web3-inspired autonomous systems and entities. McConaghy [35] is among the earliest to propose the concept of a web3-based self-governing and self-sustaining entity. He envisions a future where the integration of AI with blockchain and DAO creates smart and decentralized self-operated physical world objects such as self-own cars and self-own infrastructure such as highway roads and power grids, etc. Additionally, Wang et al. [8] also developed the concept of engineering ownership, which proposed the idea of a blockchain-based system with automation capabilities with distributed ownership and power shared between autonomous human or machine agents in the building infrastructure domain. Furthermore, in their works, Chang and Joha [10] also introduced the concept of civic and autonomous infrastructure to enhance inclusivity and eliminate the need for centralized ownership and control. They highlight the need to shift from centralized intermediaries, such as traditional financial institutions, to decentralized networks that promote shared governance. This shift is critical for developing infrastructures that are not only self-managed but also self-owned, operating without reliance on any centralized authority by leveraging distributed ledger technology.

Studies on the application of DAO governance have shown that human and machine agents can collaboratively create a decentralized organizational system that operates autonomously with self-ownership and without centralized control. DAO-powered entities encompass a range of applications, including a self-sustaining, generative AI-powered digital organization [27], [28], decentralized space [11] to an on-chain, and community-governed city [36]. One of the earliest concepts of the web3-based self-governing and self-sustaining entity was first introduced by Seidler et al. [37] in 2016, where they proposed the idea of a self-governing forest that can intelligently exploit its resources by selling its logging license with blockchain's smart contract automation process. In addition, the CityDAO project [36] aims to build a future blockchain-based city by leveraging land tokenization and decentralized governance. Users can purchase certificates to gain citizenship of the land through blockchain tokens to become CityDAO citizens and participate in the governance of the land, such as its policies, regulations, and future developments through the DAO governance.

In the context of building infrastructure, Ye et al. [38] have presented a DAO-enabled autonomous building maintenance system where the building operates independently according to predefined rules in smart contracts. The authors gave an exemplary use case where the embedded sensors detect equipment failures and report the issue to the DAO. The DAO then executes the appropriate responses, such as notifying the maintenance team and procuring replacement parts based on the programmed smart contract rules. The research on autonomous entities was further advanced in 2022 with the introduction of the No1s1 project, [11] where the authors have developed a prototype of a self-governing meditation space based on blockchain technology. This application demonstrates how DAO can be operated on a physical infrastructure, highlighting the potential for self-operating and self-sustaining physical spaces through smart contract-driven autonomy. However, the prototype lacks the integration of artificial intelligence, which could potentially enhance the autonomy adaptability, decision-making capabilities, and overall intelligence in managing and operating the autonomous systems. Furthermore, Dounas et al. [39] proposed a digital twin and blockchain-integrated system for the AEC industry. This framework also focuses on the utilization of DAO for collaboration between humans and computing agents in future physical infrastructure. Ly et al. [40] further advanced this concept by developing another framework that integrates digital twin and DAO applications to enable data-driven and decentralized governance in smart building facilities management. However, there is a lack of comprehensive empirical investigations into the practical application of this concept in the actual building environment. Therefore, there is a need for further investigation into the integration of AI, digital twins, and DAO alongside its empirical studies within the realm of autonomous building infrastructure.

## 3. Research Methodology

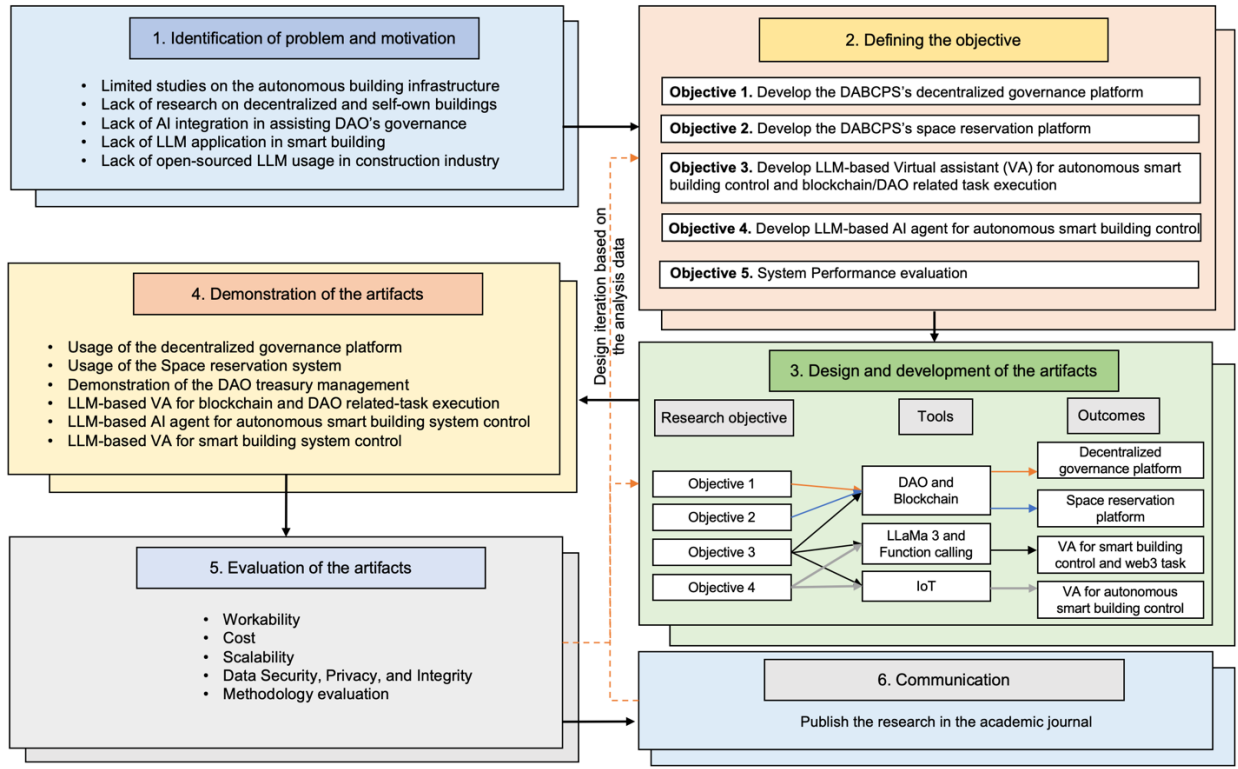

Fig. 2. Design Science Research-driven research flow

The development of the decentralized autonomous building cyber-physical system (DAB-CPS) adopted the Design Science Research (DSR) methodology, a rigorous problem-solving paradigm that emphasizes the iterative process of creating and evaluating innovative artifacts to solve identified problems in real-world contexts [41]. The DSR methodology has been widely adopted by researchers in developing blockchain-based solutions. For instance, Xu et al. [42] utilized the DSR approach to develop a blockchain-based framework aimed at facilitating decentralized carbon management, particularly in the certification of construction materials and products. Similarly, Cheng et al. [43] used DSR to create a blockchain framework for construction cost management, while Elghaish et al. [44] utilized the same methodology for the development of an AI virtual assistant for managing BIM data. The DSR process involves six different stages: (1) Problem identification and motivation, (2) Objective definition, (3) Design and development of the artifact, (4) demonstration of the artifact, (5) evaluation of the artifact, and (6) communicating the results [45]. Fig. 2 presents the development process of the decentralized autonomous building cyber-physical system with the six stages of DSR as follows:

(1) Problem identification and motivation. The literature review conducted in Section 2 reveals several research gaps, including limited studies on autonomous building infrastructure, lack of research on decentralized governance of building operation and self-owned buildings, lack of AI integration in assisting DAO governance, and the lack of open-source LLM application in the construction industry especially in the smart building domain.

(2) Objective definition. To address these knowledge gaps, this paper aims to design and develop a decentralized autonomous building cyber-physical system (DAB-CPS) framework that integrates decentralized autonomous organizations and large language models to create a smart and autonomous building infrastructure with AI-enhanced building operation and system control and financially autonomous physical spaces with distributed governance mechanism through DAO-based decentralized governance platform.

(3) Design and development. The main goal of this study, the development of DAB-CPS, will be achieved through the following objectives. (i) To develop the DAO decentralized governance platform for the operational and treasury management of the physical space (ii) To develop the physical space reservation platform for the users (iii) To create the LLM-based Virtual Assistant (VA) for autonomous smart building control and blockchain or DAO related task execution (iv) Create the LLM-based AI agent for autonomous smart building control, and conduct system performance evaluation (v) To conduct system performance evaluation. Each of these objectives will be discussed in further detail in section 4.

(4) Demonstration. The developed system mentioned above will be deployed in the actual physical building space to create a prototype of the DAB-CPS framework. Different scenarios of the system capabilities will be demonstrated, including the management of the decentralized governance platform, usage of the space reservation system, and the demo of the AI virtual assistant for blockchain and DAO-related-task execution, as well as the AI agent for smart building system control.

(5) Evaluation. A real case study in a smart building will be provided to simulate the proposed system in a realistic setting. The evaluations of the DAB-CPS framework are presented in Section 5.

(6) Communication. The design and development methodology of the proposed DAB-CPS system and prototype, as well as the evaluation results, will be published in international academic journals.

## 4. Proposed DAB-CPS framework
### 4.1. Framework overview

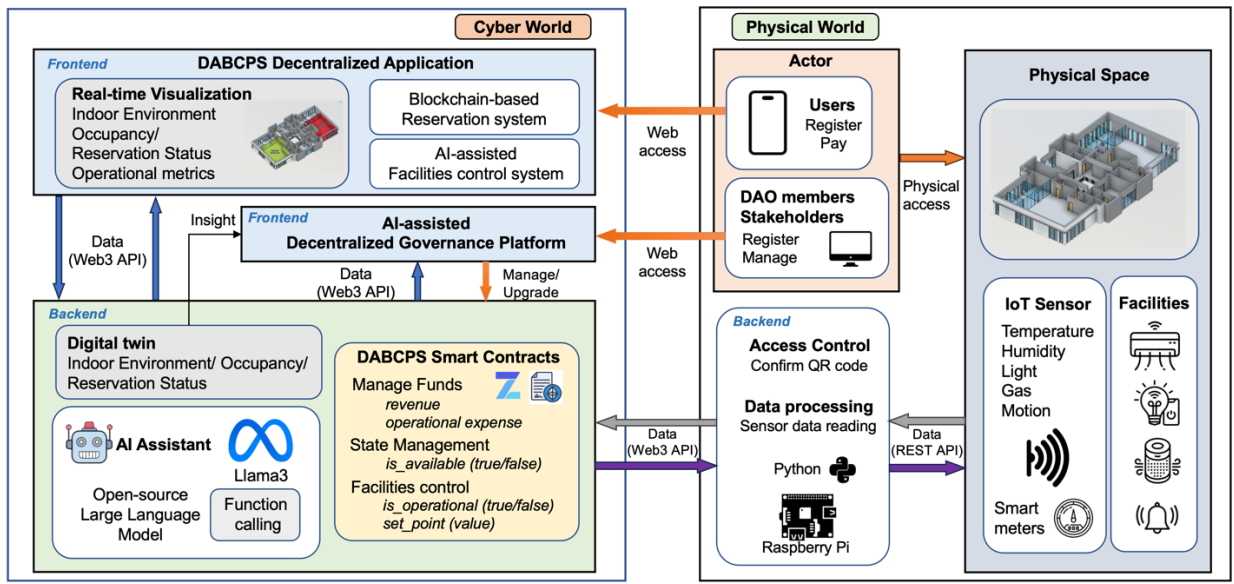

Fig. 3. Proposed Decentralized Autonomous Building Cyber-Physical System (Orange arrow: Human interaction on the CPS, Grey arrow: Data from Physical to Cyber system, purple arrow: Response from Cyber to Physical system)

In this study, we propose the concept of a decentralized autonomous building cyber-physical system (DAB-CPS), which is referred to as a web3 and AI-enabled, community-governed building infrastructure that functions autonomously with blockchain-based self-management protocols. The proposed framework aims to create a self-governing, autonomous building infrastructure leveraging blockchain technology, DAOs, and LLMs model-powered building automation systems. The DAB-CPS framework will be demonstrated through the development of a decentralized rental space. This decentralized space will be designed to earn blockchain's cryptocurrency or tokens (revenue) in exchange for renting/leasing itself to the users/occupants. The DAO's governance platform will be created to oversee the space's functional operation and financial aspect, including managing its revenue and necessary operational expenses (e.g., maintenance fee, electricity usage, etc.) with DAO treasury and blockchain-based transactions to sustain the system operation.

The architecture of this framework and the relationship between different components is also shown in Fig. 3. As depicted, the DAB-CPS framework is divided into two main domains: the Cyber world and the Physical world. The framework's cyber elements comprise different technologies, including blockchain, DAO, AI virtual assistant, and digital twins, while the physical element of the framework includes IoT devices, sensors, physical space, equipment, and human actors. In the physical world, the IoT device and sensors continuously monitor environmental conditions, while Raspberry Pi serves as the intermediary between the physical sensors and the cyber system by processing data and transmitting it to the cyber component. These sensor data are then used for digital twin visualization before feeding into the LLM-based AI agent for autonomous building operations management. The threshold values and automation logic of the building facilities are encoded in the blockchain smart contracts, which are managed by the decentralized governance platform. The AI virtual assistant also enhances user interactions with the physical space and supports DAO members in their governance activities.

The orange arrows in Fig. 3 represent human interactions with the cyber and physical system, showing how users and DAO members interact with the frontend applications and physical space. The Grey arrows indicate data flow from the Physical World to the Cyber World, including the sensor data transmission via Web3 and REST API. Purple arrows show the system's responses from the Cyber World to the Physical World, such as controlling facilities based on the command and decision from the DAO stakeholders and AI agents. Lastly, the blue arrow demonstrated the communication between the frontend and backend components of the Dapp. In addition, as illustrated in Fig. 4, the technical stacks of the DAB-CPS architecture are comprised of six layers. The user layers represent the human actors interacting with the system, which includes both DAO members who participate in governance and regular occupants who use the physical space. The main component of the application layers is the full stack Dapp, which contains key components from different layers, including the decentralized governance platform, space reservations portal, digital building twin visualization portal, and the AI-assistant portal. A Dapp with an interactive graphical user interface will be developed to provide access for the users to rental/booking services, facility control, and live digital twin visualization of the space, as well as offer the stakeholder an interactive platform for the collective governance of the physical space. Furthermore, the blockchain layer serves as the foundation of decentralized and trustless operations of the systems by encompassing the DAO-based decentralized governance platform and the blockchain-based space reservation portal. The decentralized governance platform will enable stakeholders (DAO members) to collectively manage the building operation and policy through DAO's decentralized governance mechanism, such as proposal submission, token-based voting, treasury management, and more. Furthermore, the space reservation portal allows users to reserve and cancel the space through blockchain transactions.

Moreover, the artificial intelligence layer leverages LLMs to enhance system capabilities. It includes smart assistance for decentralized governance, blockchain transactions, human-building interaction through a virtual assistant, and LLM-based autonomous building facilities automation. Furthermore, the physical infrastructure layer comprises the tangible components of the system, including the physical infrastructure, building systems and facilities, IoT edge devices, and environmental sensors, which are mainly responsible for data collection and operation of the physical space. In addition, the Data Transmission and Visualization Layer handles the processing and visualizing of the data collected from the physical layer. Its main component is the digital building twin portal, which gives the user real-time visualization of physical space conditions, including indoor environment, occupancy, and reservation status. The following sections will discuss each of these components in further detail.

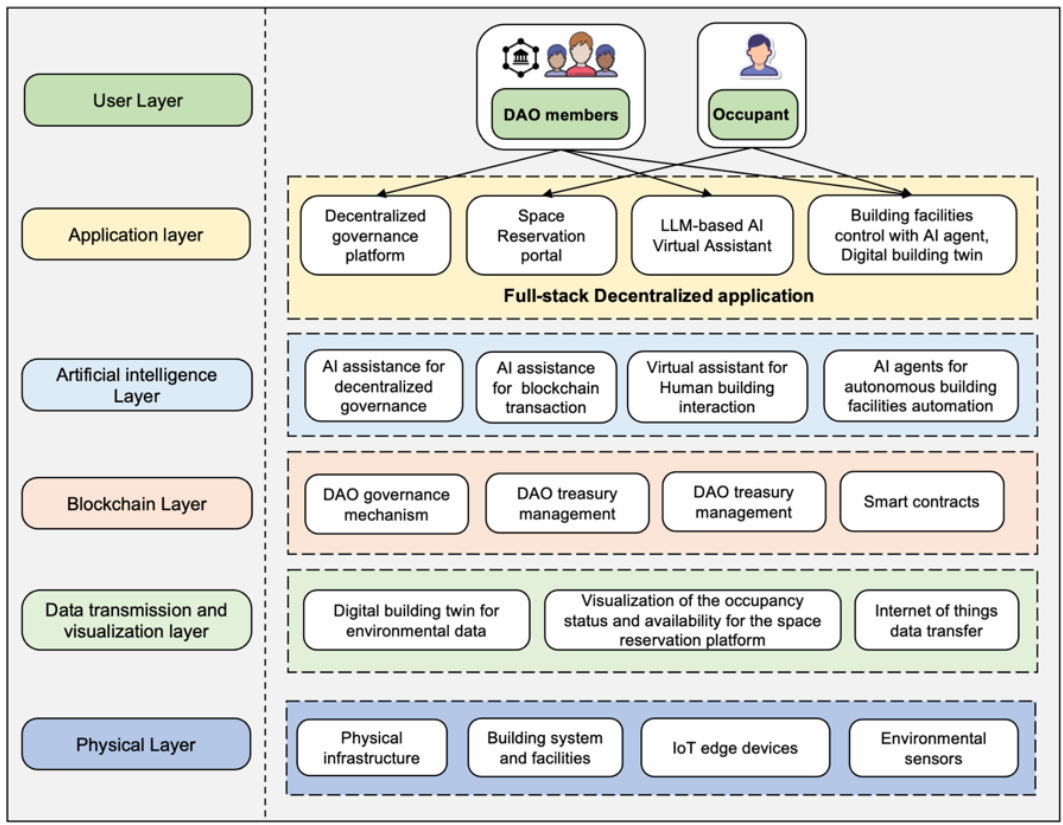

Fig. 4. Technical stacks of the Decentralized Autonomous Building Cyber-Physical System

### 4.2. Decentralized governance platform

The primary component of the proposed DAB-CPS framework is the decentralized governance platform, which serves as the central decision-making and operational hub of the physical space. Fig. 5 illustrates the architecture framework of the platform's framework, including its DAO-based governance mechanism and main functionality. The main objective of this decentralized governance platform is to facilitate transparent, secure, and decentralized governance processes among key stakeholders to collectively manage the physical space and building infrastructure, including facilities management tasks such as building facilities automation logic, operation, and maintenance issues as well as financial related matter such as income, budget allocations, expense for the physical space with DAO treasury and blockchain-based transactions, ultimately enabling a self-sustaining, nonprofit-seeking operation. Depending on the ownership structure of the physical space (public or private), DAO members can be comprised of various stakeholders, including building owners, facility managers, maintenance team, or broader occupant demographics who will be responsible for data-driven and collective governance of the DAO physical space by collectively propose and vote on different matters such as funding allocation, space operational settings, maintenance schedules, etc.

Voting mechanisms are essential to the operation of the decentralized governance platform. Upon the initial deployment of the DAO on the blockchain network, a predetermined quantity of governance tokens is minted and allocated to members based on their specific roles and responsibilities within the organization. Key members will receive a larger allocation of governance tokens, which grants them more voting power and governance rights compared to regular members. The governance tokens used in this study are based on the ERC-20 standard [46] as they provide token fungibility and transferability among DAO members, which facilitate the fractionalized ownership of the DAO system and the delegation of voting power. The platform will implement a token-based quorum voting mechanism [47], where the voting power will be proportionate to the number of tokens held by the DAO members. The DAO's governance process encompasses several key stages, including proposal submission, vote casting, proposal queuing, and execution. These procedures are governed by predefined conditions embedded within the DAO's smart contract, established during its initial deployment. These conditions include voter eligibility criteria, available voting options, duration of voting periods, implementation of voting delays, and quorum requirements. Upon the submission of a proposal for on-chain voting, a mandatory delay period is initiated. This interval allows voters to thoroughly examine the proposal, engage in discussions, and formulate informed decisions before casting their votes. Once the active voting period starts, eligible members can securely sign and cast their votes, with results transparently recorded on the blockchain. If a proposal achieves the required threshold, a minimum delay period is enforced before smart contract execution. This pause provides an opportunity for members to contest decisions or retract support if necessary, ensuring thoughtful consideration and mitigating hasty actions. Following this delay, the smart contract autonomously triggers the corresponding actions and updates the blockchain ledger. These actions may encompass fund transfers, contract modifications, or the initiation of any governance activities such as access control, incentivization, and more.

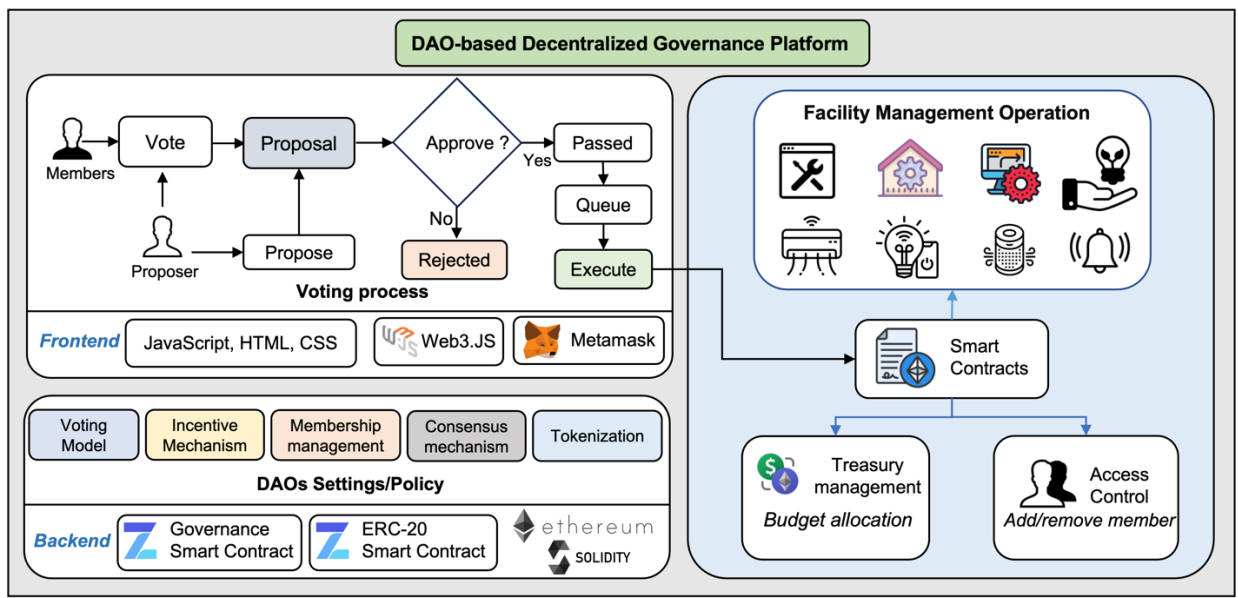

Fig. 5. The framework of the decentralized governance platform.

### 4.3. Space reservation portal

The Space Reservation Portal functions as a vital component of the decentralized autonomous building system, serving as the primary income source for the proposed DAB-CPS system. The portal provides an intuitive, web-based interface that allows users to easily navigate available time slots, make reservations, and manage their bookings. Users' booking and cancellation will be facilitated by the blockchain's smart contracts. This ensures that all transactions are immutable, traceable, and executed automatically without intermediaries. When a user makes a booking, the associated funds are automatically transferred to the DAO's treasury through smart contracts.

## 4.4. AI virtual assistant portal

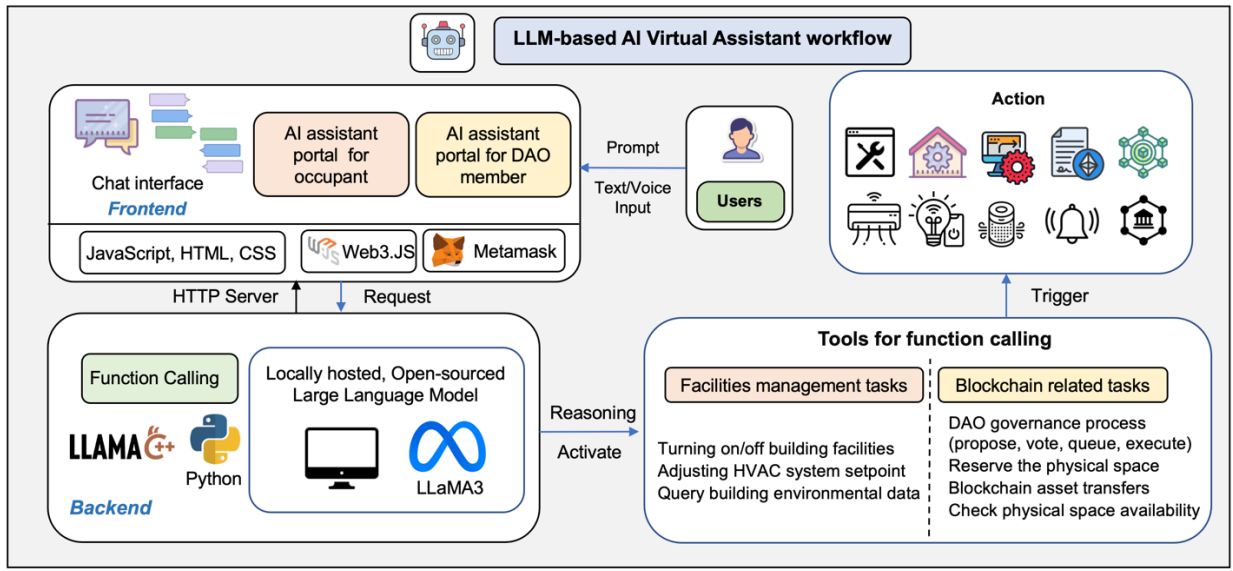

Fig. 6. Workflow of the Virtual Assistant

The AI Virtual Assistant portal aims to facilitate the human-building interaction aspect within the proposed DAB-CPS framework. This portal aims to provide smart and personalized assistance to both the DAO members and regular users for interaction with the physical building facilities and the Dapp components such as the decentralized governance platform and space reservation portal. For regular users, the AI virtual assistant serves as an interface for interacting with the physical space. Users can communicate with the AI virtual assistant through text and voice input to control various building facilities, adjust setpoints for the specific building smart facilities, or turn systems on or off as needed. The assistant also provides real-time information on indoor environmental conditions by accessing live sensor data reading from the IoT device. Additionally, it integrates with the blockchain-based reservation system, allowing users to check room availability and book spaces directly through the AI interface. The AI virtual assistant also facilitates the governance tasks of the DAO members and administrators of the physical space. The AI virtual assistant also supports the DAO members or owners in their interactions with the blockchain. DAO members can instruct the AI virtual assistant to vote on their behalf by calling the appropriate smart contract functions for executing transactions, including voting, queuing and exacting proposals, and transfer of blockchain assets and tokens. It can also facilitate the creation and submission of new DAO proposals, thereby streamlining the governance workflow.

Central to the AI virtual assistant is a locally hosted, open-source LLM and its function calling capabilities, specifically the LLaMA3 model by Meta [48]. The use of local and open-source LLM is driven by several factors, including enhanced data privacy, and reduced operational costs. By keeping all interactions and data processing within the local infrastructure, the system also ensures independence from third-party entities and maintains strict control over sensitive information. The AI virtual assistant Portal workflow is illustrated in Fig. 6, which demonstrates the integration of frontend and backend components. Users interact with the system through a web-based chat interface. The submitted prompts by the user are transferred to the backend and processed by the local LLM model. Upon receiving a query, the LLMs model leverages its reasoning and function-calling capabilities to understand the user's request and activates the appropriate predefined tools for either facilities management tasks or blockchain-related operations before executing the corresponding tasks. In this study, the quantization of the LLM is performed using the Llama.cpp library [49]. Quantization is an essential operation for the reduction of the model's size and computational requirements, which is particularly important for deploying the AI virtual assistant on local hardware [50].

### 4.5. Digital building twin

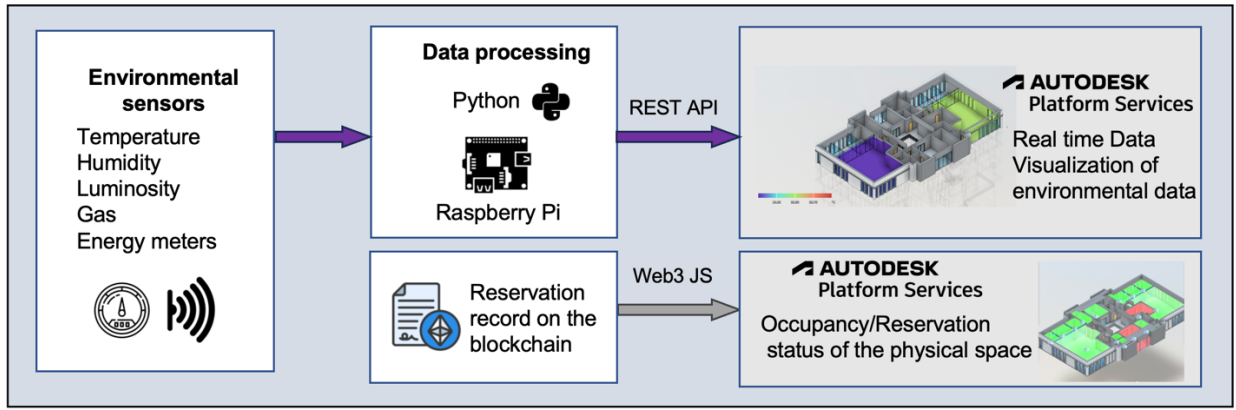

Fig. 7 Workflow of the Virtual Assistant

The digital building twin offers a comprehensive virtual representation of the physical structure within the proposed DAB-CPS framework. This advanced digital replica enables real-time monitoring of various building performance metrics, including energy usage, environmental data, and space availability. The twin's architecture incorporates both static and dynamic data components, with the static elements derived from a detailed Building Information Modeling (BIM) model. There are two different digital twin visualizations available in this study (Fig. 7). The first digital twin visualization aims to provide stakeholders with valuable insights into the building's environmental condition and corresponding energy performance, which facilitates data-driven decisions for the physical space facilities management strategies. The real-time and historical environmental data such as humidity level, temperature, light intensity, carbon monoxide gas, and energy usage information are collected with and processed by edge devices before being transmitted to the first digital twin platform. Furthermore, the second digital twin platform focuses on the dynamic representation of the physical space's availability and reservation status.

### 4.6. LLM-based autonomous building operation.

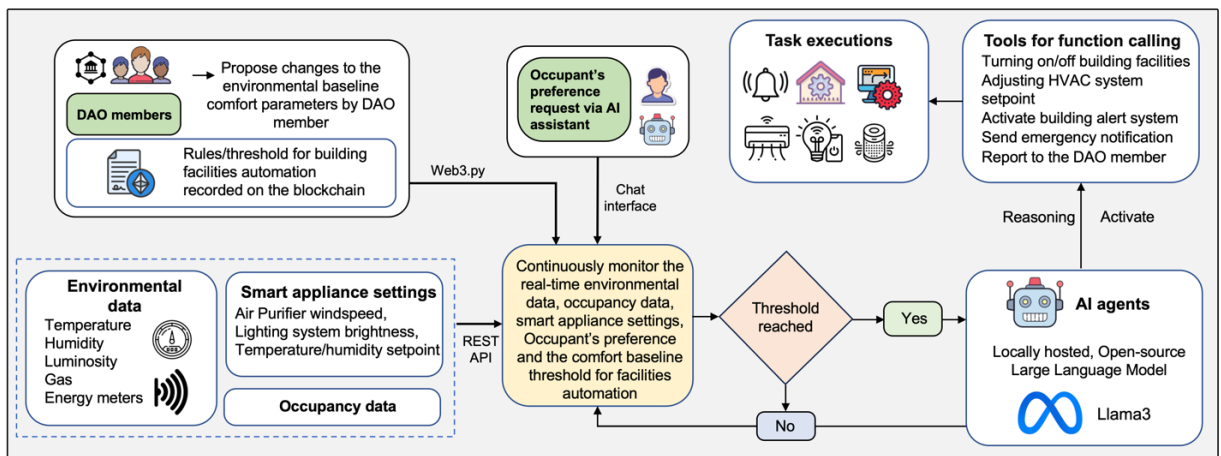

Fig. 8. Summary of the LLM-based AI agent for autonomous building operation.

In addition to the AI virtual assistant in the previous section, the DAB-CPS framework also leverages LLMs and function callings as an AI agent to oversee the autonomous building operation and facilities management. The proposed LLM-based autonomous building operation framework leverages the integration of IoT devices and sensors, smart building facilities such as the HVAC and lighting system, LLMs, and blockchain technologies to create decentralized and automated control of building systems (Fig. 8).

An array of environmental sensors and IoT Devices will be used to collect environmental data such as temperature, humidity, luminosity, carbon monoxide gas levels, as well as the energy usage information and occupancy level within the physical space. Raspberry Pi devices are used to process these data and feed them into the AI agent through REST API. In addition, the threshold parameters used for the automated smart building facilities operation (e.g., max or min temperature, humidity, etc.) will be extracted from the blockchain smart contracts, ensuring secure and transparent operations. These threshold values are the baseline comfort parameters for ensuring optimal comfort and indoor environmental quality and are defined by DAO members, who serve as administrators for the physical spaces. These thresholds are recorded on the blockchain for transparency and immutability and form the basis for the AI's decision-making process. During the operation, the LLM-based AI agent will continuously compare real-time sensor data from the physical space against these predefined thresholds on the blockchain. When the environmental data values exceed these thresholds, the AI agent utilizes its function-calling capabilities to control various building systems, including smart lighting and HVAC systems. These operations involve adjusting setpoints, activating or deactivating devices, or triggering alerts to maintain optimal building conditions.

In addition to the rule-based automation, the LLM-based AI agent is also designed to perform contextual decision-making, which allows for more sophisticated and dynamic system control. The proposed AI agent can interpret multiple variables simultaneously, such as the current settings of smart appliances and the occupancy levels within the building. In this study, the occupancy data will be periodically analyzed, and the AI agent will adjust the performance of building systems (e.g., HVAC performance or lighting intensity) based on occupancy level to dynamically optimize energy consumption. In addition to occupancy-based adjustments, the AI agent can also process natural language inputs from users, such as feedback or specific requests, and dynamically modify its autonomous control logic in response. This enables the system to accommodate immediate individual preferences or situational changes. The proposed AI agent provides a level of flexibility and adaptability that surpasses traditional automation systems, making the building's operational framework not only more efficient but also more responsive to dynamic environmental and user-driven changes.

## 5. Proof of concept: Implementation and validation of the prototype

In this section, a case study with the developed prototypes is used to validate the viability and functionality of the DAB-CPS framework. The code for the technical implementation of the DA-CPS prototypes is available under an open-source license [51]. The tools, coding languages, and development environments employed for each module of the DAB-CPS prototypes are summarized in Table 1.

Table 1. Tools used for prototype development.

| Tasks | Programming language (packages) | Development environment |
|---|---|---|
| Frontend web pages development | React JS | Visual Studio Code |
| Smart contract development | Solidity | Brownie |
| Digital building twin | JavaScript (Autodesk API) | Visual Studio Code |
| IoT sensors and smart home device | Python | Visual Studio Code |
| Interaction between Dapp and smart contract | JavaScript (web3.js API) | Visual Studio Code |
| AI agent's function calling | Python (llama-cpp-agent) | Visual Studio Code |

| Large language models deployment | C++ (llamacpp) | Visual Studio Code |
| Large language models Inference | Python (llamacpp-python) | Visual Studio Code |

## 5.1. Development of the Dapp backend
### 5.1.1. Smart contract design and development

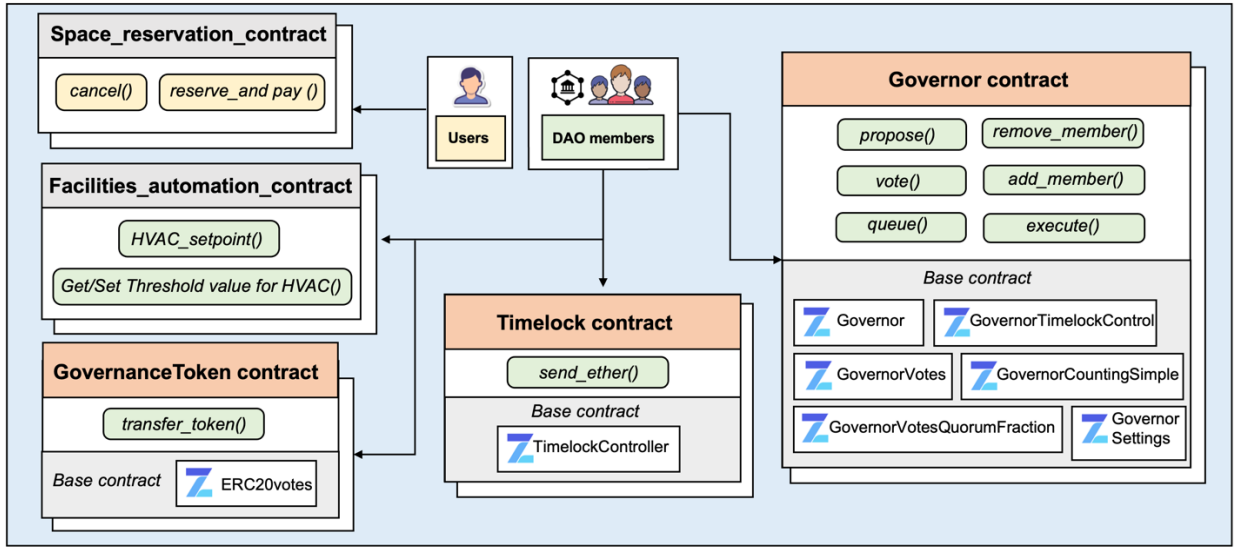

Fig. 9. Design of the decentralized governance platform's smart contracts

This section details the design and development of smart contracts for the backend of the proposed DAB-CPS framework, including Dapp components such as a decentralized governance platform, space reservation system, and the automation logic of building facilities operation. The five smart contracts in this prototype, the DAO Governor contract, Governance Token contract, Timelock contract, space reservation contract, and facilities automation contract, were written in the Solidity programming languages [52] and compiled using Brownie library, a python-based development environment that allows developers to write, test, and deploy Ethereum smart contracts. Given the collaborative aspect of the governance platform, a public/permissionless blockchain network is chosen for the design and development of this DAO application. Ethereum blockchain [53] is chosen in this study as it provides a diverse set of tools for DAO development. For research purposes, Ethereum's test network, Sepolia, will be used as it provides all the necessary functions of Ethereum without any fees.  The design of the five smart contracts and the relationship between their function and the roles of actors in the proposed framework are illustrated in Fig. 9.  The decentralized governance platform is structured around three primary smart contracts: the DAO Governor contract, the Time Lock contract, and the Governance Tokens contract. This study utilized the base smart contracts from the OpenZepplin library [54] , including the DAO governor contracts and ERC-20 contracts, as it provides secured and audited codes with community-standardized contracts for DAO development.

The DAO Governor contract Is primarily responsible for the core governance processes within the decentralized governance platform. It manages proposal submissions, voting procedures, and the execution of approved decisions. It handles the votes counting mechanism and determines the outcome of proposals based on established quorum and majority requirements. The DAO Governor contract in this study is built upon several OpenZepplin base contracts. For example, the Governor base contract provides essential functions for proposal creation and execution. The GovernorTimelockControl contract adds a security layer by introducing a delay in proposal execution. The GovernorVotes contract ensures voting power is based on the ERC-20 governance tokens [46]. GovernorCountingSimple implements a straightforward vote-counting mechanism, while GovernorVotesQuorumFraction enforces a quorum based on a fraction of the total token supply. Additionally, GovernorSettings allows for the configuration of governance parameters such as voting delays and periods. The DAO Governor contract works closely with the Timelock contract to schedule the execution of approved proposals and interfaces with the Governance Tokens contract to verify member voting power for proposals.

In addition, the Governance Token contract manages the distribution and administration of the platform's fungible tokens, which represent voting power within the DAO-based governance framework. This contract handles the minting of new governance tokens upon deployment. It functions as the medium for both governance and transactional activities within the system. The Governance Tokens contract inherits from the ERC20Votes base contract, which extends the standard ERC20 token with voting and delegation capabilities. This enables token holders to vote directly or transfer their voting power to other DAO members. In this platform, it facilitates the delegation of voting power by allowing the allocation and transfer of tokens between members. This process is handled by the transfer_token function within this contract. The Governor Token contract also interfaces with the DAO Governor contract to determine voting weights and eligibility for proposal submissions.

Additionally, the Timelock contract serves as a security measure, enforcing governance decisions by introducing a mandatory delay between the approval of a proposal and its execution. This contract is extended from the TimelockController base contract from the OpenZepplin library. It queues approved proposals for a specified delay period and executes them only after this period has elapsed. This ensures that all stakeholders have adequate time to review approved decisions and make any changes if necessary. In contrast to the Governance Token contract, which controls fungible tokens, the Timelock contract manages the DAO organization's monetary assets, such as Ethereum cryptocurrency. One key function within the Timelock contract is send_ether, which facilitates monetary transactions between the DAO organization and external contract addresses. In addition, the space reservation contract contains two important functions, reserve () and cancel (), which allow the users to reserve the physical space and make corresponding cancellations on the space reservation portal. Once the user reserves the space, the rental fee will be automatically transferred to the Timelock contract of the decentralized governance platform. Finally, the facilities automation contract contains the set and get function for the threshold variables. The set threshold variable function can be called by the DAO member from the decentralized governance platform to define the threshold parameters for the smart building facilities operation. The get threshold variable function will be called through web3 JS by the LLM-based AI agent for comparison against the real-time environmental data.

In addition, upon development, the three DAO contracts (Governor contracts, Governance Token contracts, and Timelock contracts) were deployed to the Ethereum Sepolia test network. It is also crucial to highlight that, by default, the user who initially deployed the governance token contract is the owner of all that governance token contract, which contradicts the notion of decentralized governance where no individual is supposed to be solely in control in the DAO's treasury. To address this problem, this study deployed these three contracts simultaneously at once in a single Python script. The initial DAO member can also be determined by the initial allocation of ERC-20 tokens within the Governance token and a certain user address within the first deployment of the contract. By doing so, the ownership of the Governance Token contract and Timelock contract, which are crucial for the management of DAO's ERC-20 token and Ethereum cryptocurrency assets, will be successfully transferred to the Governor contacts right after the initial deployment of DAO. Therefore, future governance actions such as monetary and token assets will be monitored by the Governor's contracts, which are controlled by all DAO members instead of a single user or authority. The "onlyDAOmember" modifier is used in proposed smart contracts to enforce access control and security to important functions within DAO. Regular users or non-DAO members can only access the space reservation contract. The five smart contracts developed in this study are presented in Fig. 10.

Fig. 10. Summary of smart contract within the DAB-CPS prototypes. (a)Space reservation contract (b)Building automation contract (c)Timelock controller contract (d) DAO Governor contract (e) Governance token contract

### 5.1.2. Digital building twin

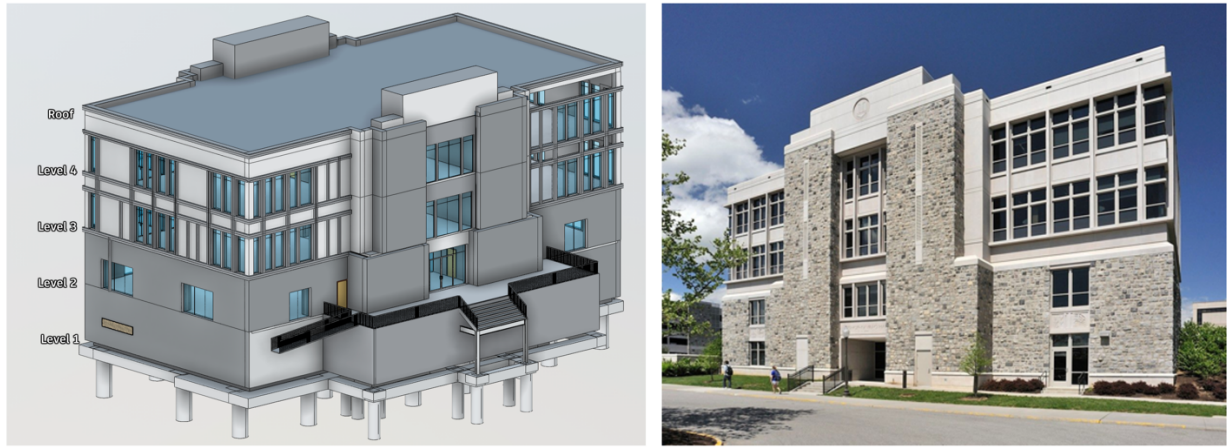

Fig. 11. Myers-Lawson School of Construction's Bishop-Favrao Hall and its BIM model

As previously mentioned in section 4.5, the DAB-CPS framework incorporates two distinct digital building twins of a single physical building. The first digital building twin aims to provide a visualization of the environmental condition and energy usage within the building, while the second digital building twin seeks to provide the reservation status of the physical space. This study chooses Bishop-Favrao Hall, home to the Department of Building Construction at Virginia Tech, as the location for the case study to implement and evaluate both digital twins and the DAB-CPS prototype. The BIM model of Bishop-Favrao Hall was developed using Autodesk Revit 2024 (Fig. 11). The environmental data for the first digital twin is collected using a network of sensors managed by a Raspberry Pi 4B single-board computer. The system interfaces with a series of environmental sensors and IoT devices to continuously monitor and collect real-time data. Key sensors used in this setup include (1) DHT11 Humidity and Temperature Sensor (2) Light Intensity Sensor (3) MQ-2 Gas Sensor and (4) Groove Smart Plug for Energy Metering. These sensors' data are read and processed on the Raspberry Pi 4B using Python libraries such as Adafruit_DHT (for the DHT11) and Rpi.GPIO, and Adafruit_MQTT. Once the environmental data is collected and processed on the Raspberry Pi 4B, it is transmitted to the first digital twin platform via a REST API, which is facilitated by Python's Flask library. The data is sent in JSON format, which includes the sensor readings for temperature, humidity, light levels, air quality, and energy consumption. For the digital twin development, the Autodesk Platform Service's Model Derivative API is utilized to integrate the dynamic environmental data into the BIM model. It allows for the real-time visualization of sensor data onto the digital building model, which allows DAO members to monitor the real-time conditions of the building. Fig. 12 presents the summary of the digital twin development workflow.

Moreover, the second digital twin platform focuses on dynamically representing the physical space's availability and reservation status. The dynamic data for this building twin is the reservation status of various spaces within the building. When a room or space is booked, the reservation is recorded on the Ethereum blockchain through a dedicated space reservation smart contract. This digital twin utilizes the Web3.js library to retrieve booking information for each space and continuously updates the vacancy status in real time. This information is transmitted to the digital twin system through the REST Web3.js library, ensuring accurate, real-time reflection of space utilization across the building.

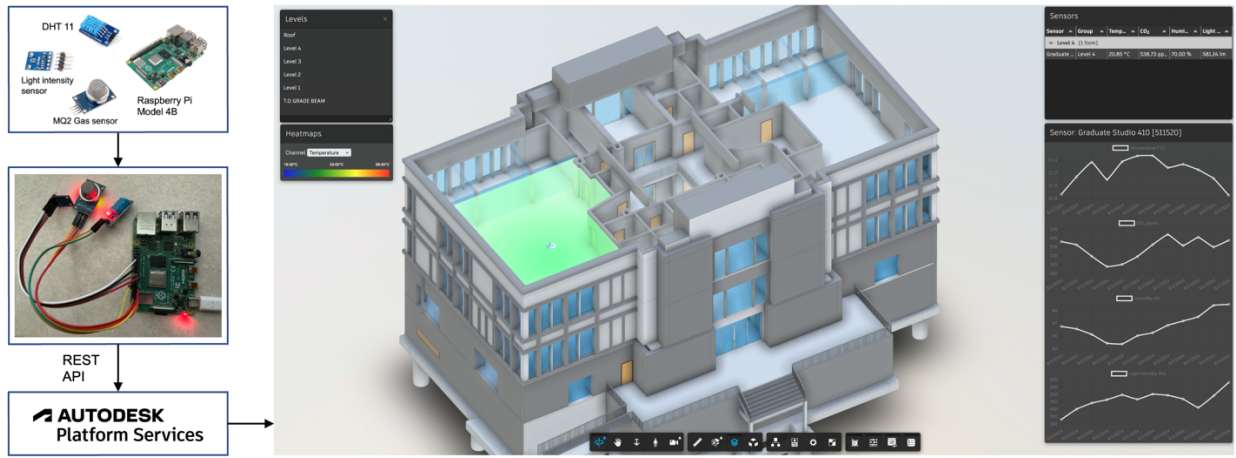
Fig. 12. Workflow of digital building twin development.

### 5.1.3. Deployment of the AI agent and virtual assistant module

The proposed AI agent and virtual assistant in this study is powered by LLaMA 3 8b, an open-source LLM with eight billion parameters developed by Meta [55]. The number of parameters of a model directly affects the accuracy, responsiveness, and functionality of the AI system. The model with a high number of parameters tends to provide better output quality, improving both the general language understanding and the precision of function calling, which is a critical feature in this AI system. The workstation used for the LLaMA4 model deployment in this study is an Apple MacBook Pro with an M1 Max chip with 32GB of RAM. To run the LLaMA 3 model efficiently, this study employs a quantized version of the model using the llama.cpp library. Llama.cpp is a tool that allows the execution of quantized LLMs on local hardware with support for different types of GPU. One key advantage of this tool is its ability to reduce the size of the model through quantization, which involves representing the model's weights in lower precision formats. This enables running complex models on devices with less computational power without severely compromising performance. The 8-bit quantized model of LLaMA 3 8b is used in this study.

Function calling is one of the core features of the proposed AI virtual assistant. In this context, function calling refers to the AI's ability to analyze user requests, extract key information, and invoke predefined tools or functions to perform tasks. This study leverages llama-cpp-agent, a Python-based package, to implement the function-calling capabilities of the proposed AI virtual assistant. In this study, various Python functions and tools were developed for the AI virtual assistant to handle both blockchain-related tasks and facilities management operations. These include turning building facilities on or off, adjusting the setpoints of HVAC systems, querying environmental data, sending emergency alerts, and managing decentralized governance processes such as proposing, voting, queuing, and executing DAO proposals. Additionally, the AI virtual assistant supports functions like reserving physical spaces, checking space availability, transferring blockchain assets, and signing blockchain transactions.

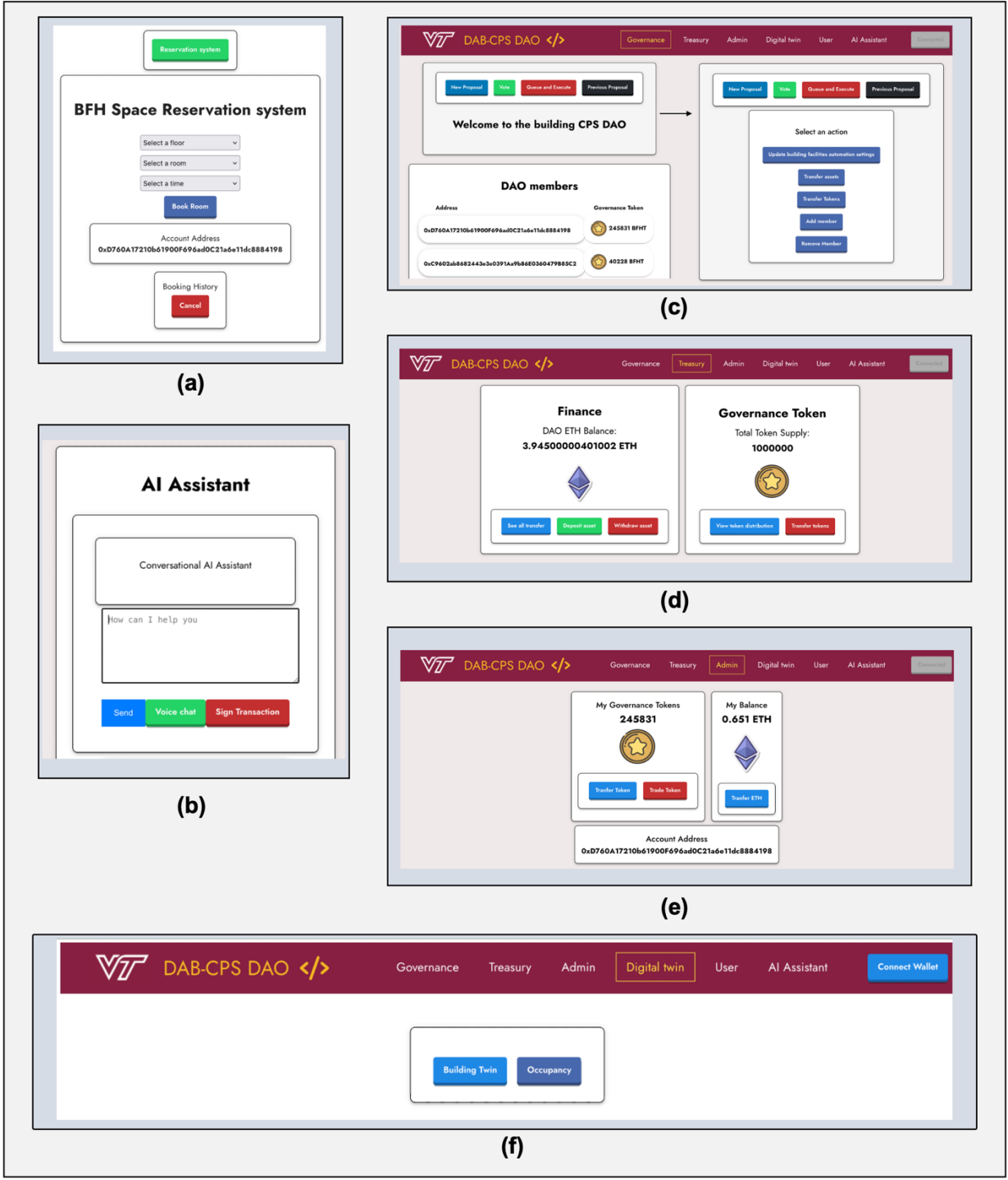

Fig. 13. Front-End of the DA BCPS Dapp: (a) Space reservation portal (b) AI assistance portal (c) Governance Portal (d) DAO Treasury tab (e) DAO Administrator tab (f) Digital twin tab.

### 5.2. Development of the Dapp Frontend

The front end of the Dapp for the proposed DAB-CPS prototype was developed using React JS due to its flexibility, modular structure, and compatibility with web3 JS, which facilitates the interaction between the Ethereum blockchain and the web application. This frontend offers an intuitive interface for both regular users and DAO members to interact with the physical space and the blockchain system. Additionally, MetaMask was integrated into the application to link the users' Ethereum wallets and accounts to the Dapp to facilitate blockchain-related transactions. As illustrated in Fig. 13, the Dapp is comprised of six main tabs: Governance, Treasury, Admin, Digital Twin, User, and AI assistant. The Governance tab allows DAO members to propose and vote to implement different building operation-related governance actions, such as adding or removing members, transferring governance tokens or cryptocurrency, and updating lighting, HVAC, and facilities automation settings. The Treasury tab displays the governance and monetary assets of the entire DAO, including governance tokens and Ethereum cryptocurrency. DAO members can propose and vote to transfer these assets from the DAO organization to specific addresses. The admin tab provides information on the governance and monetary assets of individual DAO members. Similar to the Treasury tab, DAO members can transfer their assets to other addresses through this section. The Digital Twin tab displays the two digital twins described earlier: one representing the environmental conditions and energy usage of the building, and the other showing the reservation status of the physical spaces within the building. In the User tab, regular users can access the space reservation system to book rooms and times. Finally, The AI virtual assistant tab enables both users and DAO members to interact with the AI virtual assistant. To enhance accessibility, the AI virtual assistant also integrates the Web Speech API from node JS [56], which provides speech-to-text functionality for the React JS app, allowing users to input queries or commands through voice interaction. This feature supports a more intuitive interface, enabling seamless communication between users and the AI virtual assistant.

### 5.3. Physical space and related equipment

The physical space used for this study's DAB-CPS prototype testing is meeting room at Myers Lawson school of construction. While the room is equipped with a native HVAC system and integrated lighting, it is not available on demand for public use for direct access to control these systems. To simulate the required environmental management capabilities for the study, a range of smart home devices has been installed to facilitate access control from AI virtual assistant and decentralized system on the air quality, humidity, lighting, and fan speed within the physical space (Fig. 14). To simulate air quality control within the space, this study uses a smart air purifier device, Xiaomi Smart Air Purifier 4 Compact, which is equipped with multiple fan speed configurations. This allows adjustments to airflow and purification settings within the room by the AI virtual assistant and the autonomous AI agent. Next, a smart humidifier, the Govee Smart Humidifier H7141, is used to facilitate the adjustment of humidity levels within the room. The Xiaomi Mi Smart Standing Fan 2 is also integrated into the system by offering multiple fan speeds for personalized airflow control. This device allows the AI agent to demonstrate its ability to regulate air circulation within the space, simulating traditional HVAC systems. For lighting control, this study used a smart bulb, Yeelight Smart Light Bulbs W3 as it offers adjustable brightness levels, which mimic the control of indoor lighting conditions. These smart devices were selected for their flexibility and compatibility with the research objectives, specifically, the availability of APIs and open-source tools for smart home control.

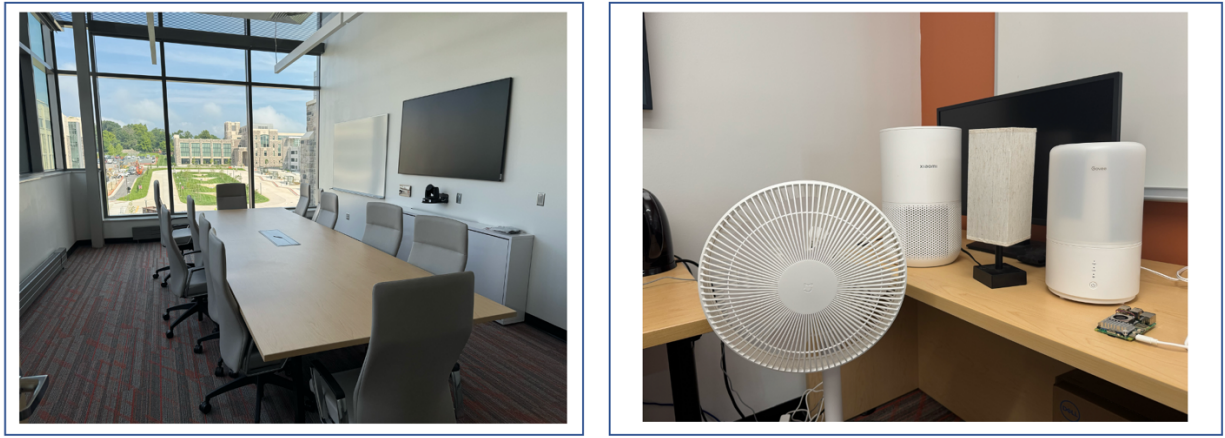

Fig. 14. Physical spaces and equipment used in the experiment. (a)Physical space (b) smart home appliance

### 5.4. Implementation

This section describes the implementation of DAB-CPS prototypes in the physical space to verify their feasibility. While the proposed DAB-CPS prototype has been fully developed with all the software and necessary environmental sensors and hardware components, the testing of user interaction still requires the involvement of key building stakeholders, including facility managers, building operators, and other decision-makers. However, since the technologies used in the DAB-CPS prototypes such as blockchain have not yet achieved mainstream adoption in the construction industry, the involving stakeholders' testing would require significant time for coordination and specialized technical expertise. Given these constraints, a scenario-based evaluation approach was adopted to simulate the user interaction aspect of the system. This approach allows the system's functionality, performance, and efficiency to be tested across various controlled scenarios without requiring the immediate participation of real stakeholders. Scenario-based validation has been widely used in different blockchain-related studies [57], [58], [59] and provides a feasible, effective method to demonstrate the viability of the technology in different practical contexts. In this study, the validation process was structured around several key scenarios, including user reservation on the space reservation portal, income and expense management by the DAO entity, proposal voting by DAO members, and AI virtual assistant demonstration for blockchain-related tasks and facility management tasks. The following sections will provide more details on each of the scenarios and its validation process.

### 5.4.1. Implementation Preparation

For the implementation of the DAB-CPS prototype and Dapp, a scenario was designed to include five distinct stakeholder roles: four facility managers and one occupant. Each of these roles was assigned an Ethereum account funded with 1 Sepolia testnet Ethereum token to participate in the blockchain operations. One of the users was designated to deploy the DAO smart contract, which included the governance contract, token contract, and Timelock contract. In this setup, the governance token, named BFHTokens, was minted in a total supply of 1,000,000 tokens. At the time of deployment, three out of the five accounts were allocated 10,000 BFHTokens each, which granting them as DAO members status. The remaining two accounts were designated for other purposes: one would be a regular user responsible for making space reservations, while the other would be added as a DAO member through a vote by the existing members. The smart devices in the physical space were configured with specific environmental setpoints to simulate real-time control and demonstrate the capabilities of the DAB-CPS system. The baseline comfort parameters were set within the following ranges: room temperature between 20°C (min) and 27°C (max), illuminance intensity between 50 lux (min) and 150 lux (max), relative humidity between 40% (min) and 100% (max), and carbon monoxide concentration between 400 ppm (min) and 1000 ppm (max). To further assess the energy consumption within the physical space, all equipment was connected to two smart plugs equipped with energy meters. These meters tracked the energy usage of the devices throughout the experiment, which was conducted over a seven-day period from September 14 to September 21, 2024. To simulate one month's worth of energy data within this two-day window, the measured energy consumption was multiplied by a factor of 4. This calculation was necessary to simulate the frequency of events like utilities expense payouts, where the DAO would distribute funds based on energy consumption data.

The interaction between facility managers and occupants with Dapp is illustrated through ten steps, which are labeled alphabetically from (a) to (j) in Fig. 15. Steps (a), (b), and (c) are related to pre-implementation activities such as roles assignments, account funding, and smart contract deployment. Steps (d) and (e) are related to occupant interaction with the reservation system and digital twin. Step (f) to (j) demonstrate the DAO member governance process for blockchain and facilities management-related tasks.

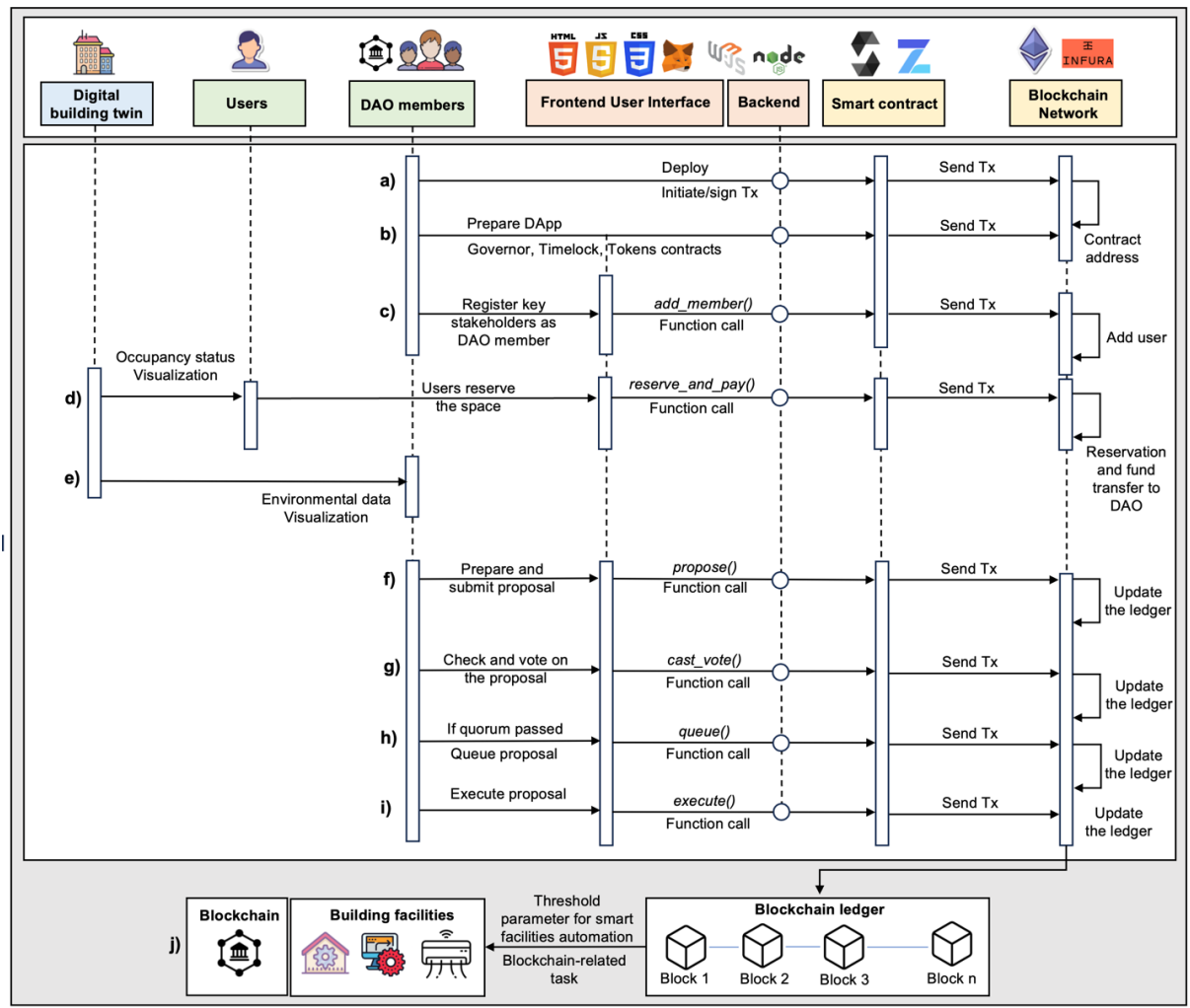

Fig.15. Sequence diagram of implementing the Dapp.

### 5.4.2. Scenario 1: User reservation on the space reservation portal

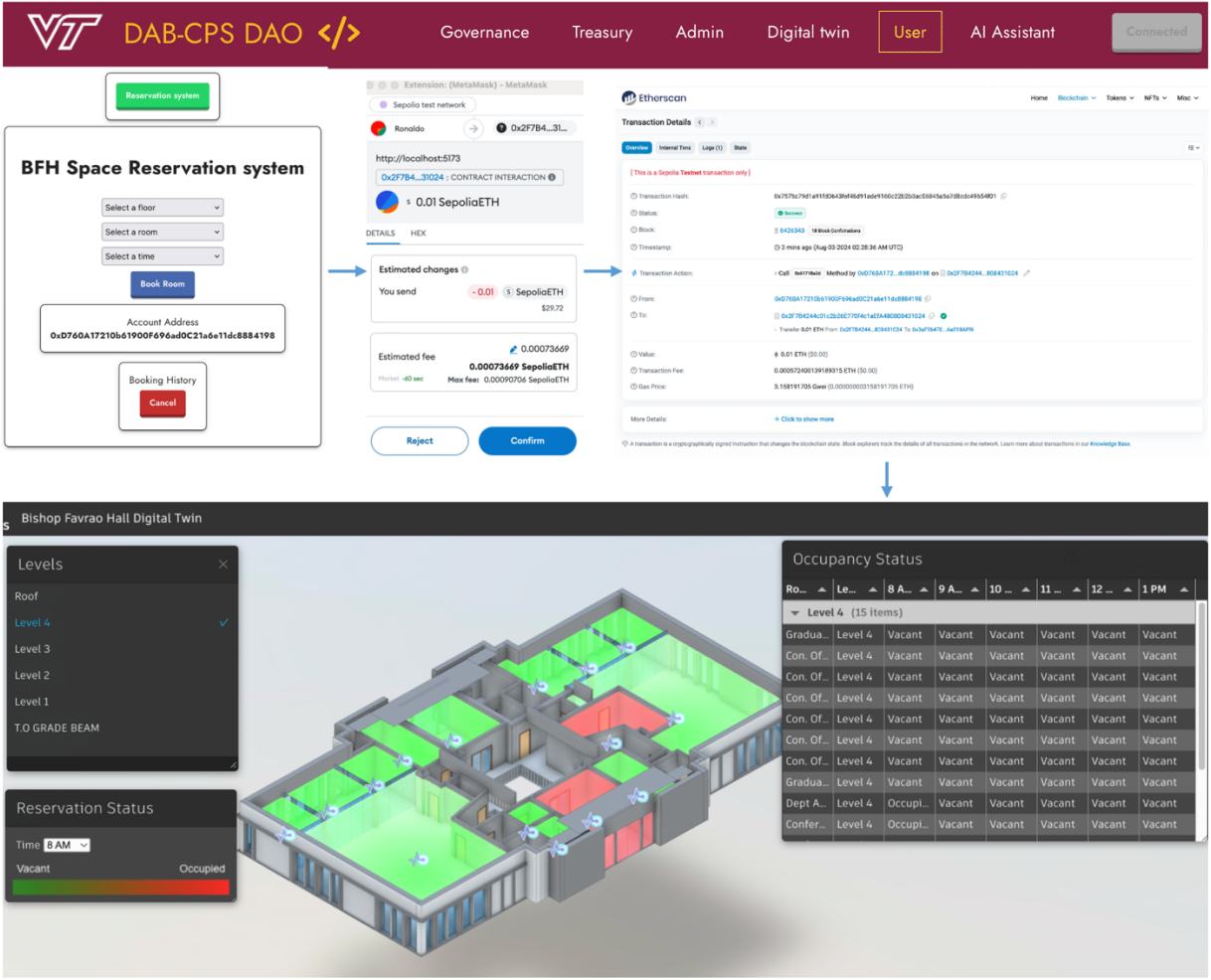

Fig. 16. Roadmap for reserving the physical space on the Dapp.

In this scenario, the main objective is to validate the user reservation process within the DAB-CPS system, demonstrating the interaction between end-users and the Dapp for booking physical spaces. As shown in Fig. 16, the user can access the decentralized space reservation portal on their phone or computer. To initiate the reservation process, users link their Ethereum wallet to the Dapp frontend via MetaMask, select a location, and specify the desired time for the reservation before booking and signing the transaction. The user pays a fee of 0.001 Ethereum for the reservation. After the booking transaction is processed, the successful booking is displayed on the portal. In addition, the space is marked as "occupied" and represented visually by a red heatmap on the building's digital twin, indicating that the room is no longer available. Additionally, the booking fees are transferred to the DAO's address, contributing to its revenue. This scenario validates the ability of users to make space reservations while simultaneously generating income for the DAO, confirming that the system functions as intended in real-world conditions.

### 5.4.3. Scenario 2: DAO Expenditure management

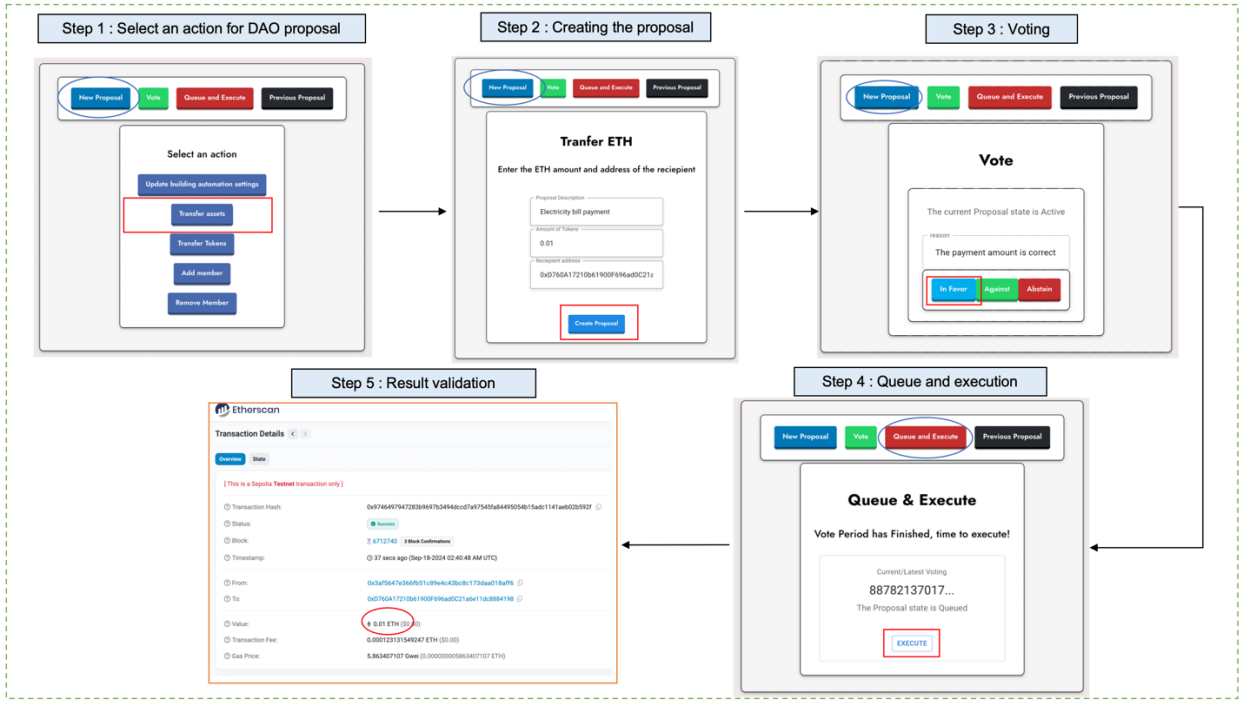

Fig. 17. The roadmap for the DAO governance process for expense management

The second scenario focuses on the validation of DAO-managed expenses, specifically demonstrating how DAO members can coordinate to manage and pay expenses related to the decentralized space. This scenario simulates the payment of electricity bills for energy usage within the space. The total energy consumption for the month is obtained via the Govee smart app, which measures energy use through connected smart plugs with energy meters. For research purposes, the electricity rate is set according to the current rate in Blacksburg, Virginia, and the total energy cost is calculated based on this rate. The equivalent cost in Ethereum is then derived by converting the dollar amount to Ethereum tokens. Next, a DAO member initiates a proposal to pay the calculated energy bill in Ethereum to the electricity provider's Ethereum address. As shown in Fig. 17, all DAO members participate in the voting process to approve or reject the expense. Once the proposal is passed, the payment transaction is executed on the blockchain. As illustrated, the Ethereum transaction can be tracked and verified through Etherscan, showing the successful transfer of the equivalent Ethereum to the electricity provider's wallet. This scenario validates the capability of the DAO to manage and approve expenses in a decentralized manner, demonstrating the system's ability to autonomously handle essential operational costs through collective decision-making.

### 5.4.4. Scenario 3: DAO Governance for updating building system operational threshold.

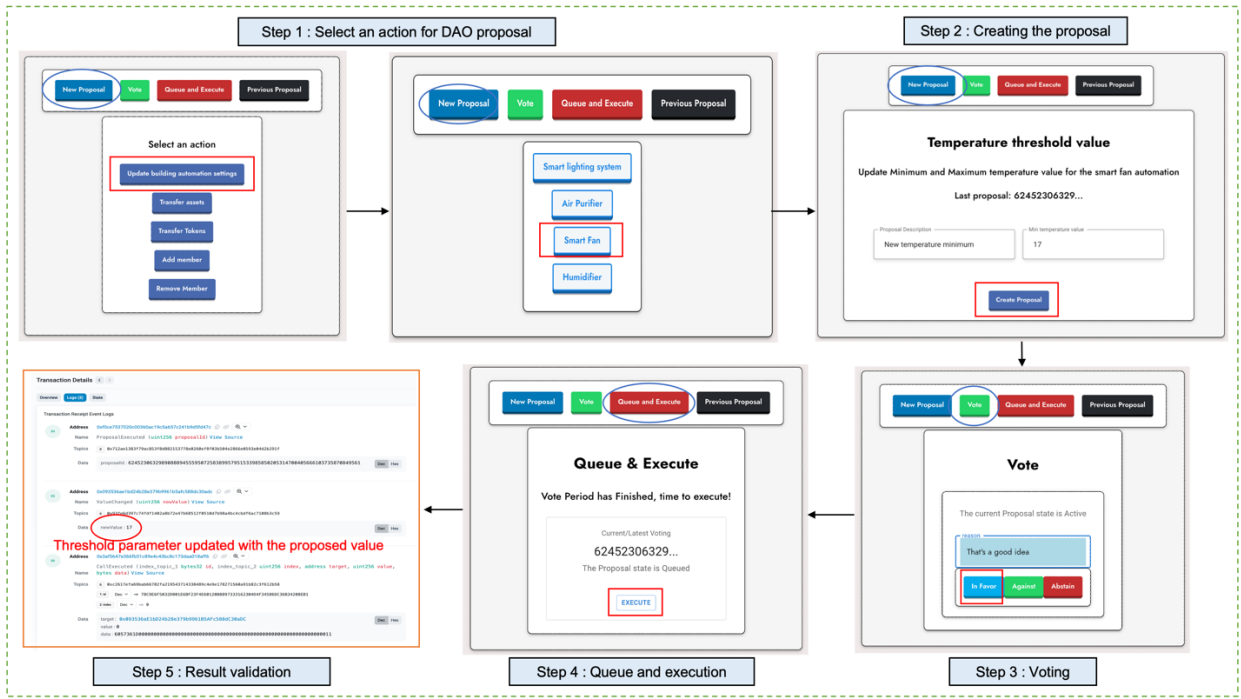

Fig. 18. The roadmap for the DAO governance process for setting operational thresholds for smart appliances.

This scenario also aims to evaluate the decentralized governance aspect of the DAB-CPS system by demonstrating the DAO's capacity to set operational thresholds for smart appliances of the physical space, such as adjusting baseline environmental comfort variables value including minimum and maximum temperature, humidity, luminance, carbon monoxide concentration level, which dictate the environmental conditions for optimal user comfort. As depicted in step 1 and step 2 in Fig. 18, a DAO member initiated a proposal to modify the minimum temperature threshold to 17 degrees Celsius. In step 3, DAO members participated in the voting process, where they could either approve or reject the proposed changes. Upon successful voting and proposal approval, DAO members queued and executed the proposal to write the new threshold values into the blockchain's smart contract (step 4). As can be seen in step 5, the updated parameters were successfully recorded on the blockchain. Finally, the DAB-CPS prototypes with its AI agent subsequently adopted these new values to autonomously control the environmental conditions in the space. Therefore, this scenario validated the DAO's ability to democratically set operational thresholds for space devices, confirming that the DAB-CPS system can effectively integrate decentralized decision-making into its autonomous building management processes.

### 5.4.5. Scenario 4: AI virtual assistant for Blockchain-Related Tasks Execution

This scenario aims to evaluate the integration of AI virtual assistant in facilitating blockchain-related tasks for DAO members and users within the DAB-CPS prototype. For validation purposes, this experiment tests the use of an AI virtual assistant to assist the Ethereum token transfers. In this scenario, the selected task involves the transfer of 0.01 Ether to the address

"0x3aF5647E366fb51C89e4c43Bc8C173dAa018AFf6". As demonstrated in Fig. 19, the user issued a command to the AI virtual assistant to transfer Ether. The prompt is then transmitted to the backend, where it is processed by a locally hosted LLMs. Upon receiving the user's query, the AI employs its function-calling and contextual understanding capabilities to interpret the request, extract useful information such as user address and amount of ether, and activate the "send Ether" function, which is pre-programmed in the list of its executable tools. The AI virtual assistant then called the Ethereum transfer function on the blockchain smart contract and prompted the front-end interface to request the user or DAO member's signature to confirm the transaction. After the successful transaction, an amount of 0.01 Ether has been sent to the address "0x3aF5647E366fb51C89e4c43Bc8C173dAa018AFf6," as requested.

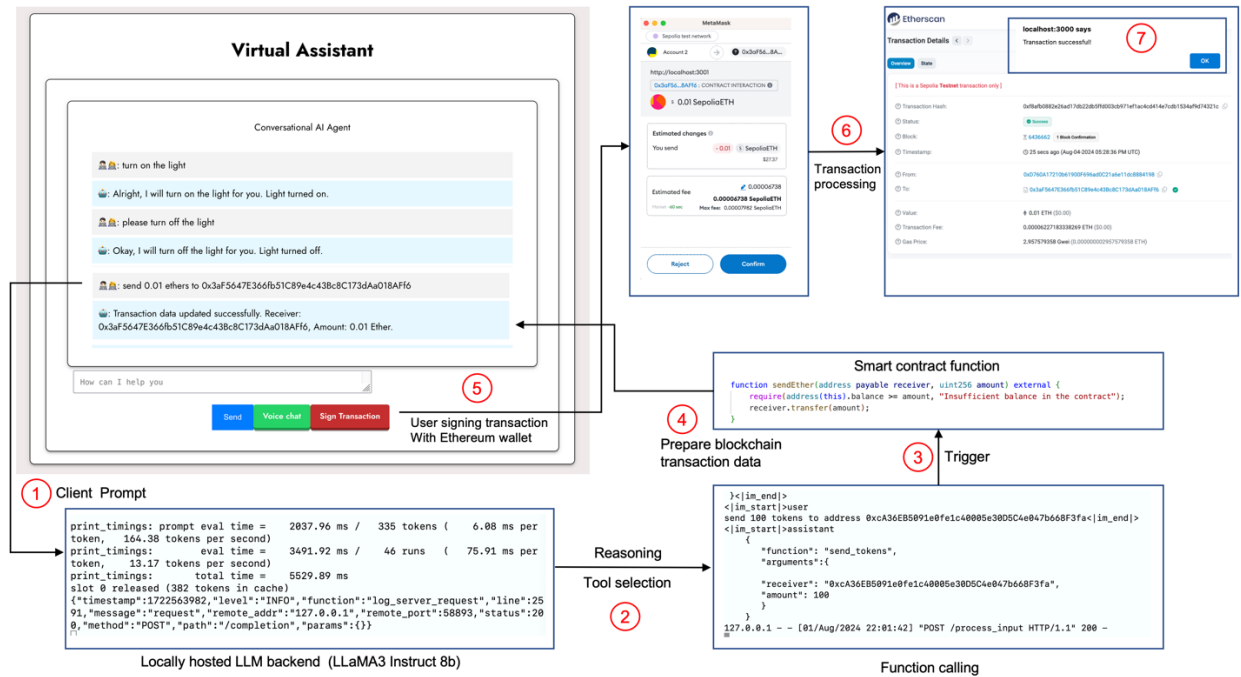

Fig. 19. Blockchain task execution using LLM-based AI virtual assistant.

### 5.4.6. Scenario 5: AI-Assisted Control of Smart Devices

This scenario aims to validate the LLM-based AI assistant's ability to control smart home devices within the DAB-CPS prototype. For this demonstration, the control of a smart light bulb was selected. Users can interact with the AI virtual assistant to issue voice commands such as turning the light on or off. Like the previous scenarios, the user's prompts are transmitted to the backend, where they are processed by the locally hosted LLMs. Upon receiving the request, the AI uses its function-calling capabilities to understand and act upon the user's instructions. It then activates a Python script that interfaces with the device's API, enabling it to control the smart light bulb as requested. In addition, instead of the simple on/off commands, users can also provide the context or hint, such as stating that the room is "too dark" or "too bright." The AI virtual assistant, leveraging its context-awareness capabilities, can autonomously adjust the brightness level of the light bulb, which can be seen in Fig. 20. This dynamic interaction demonstrates the enhanced functionality of the LLM-powered AI virtual assistant, distinguishing it from traditional AI virtual assistant that lack such nuanced environmental awareness.

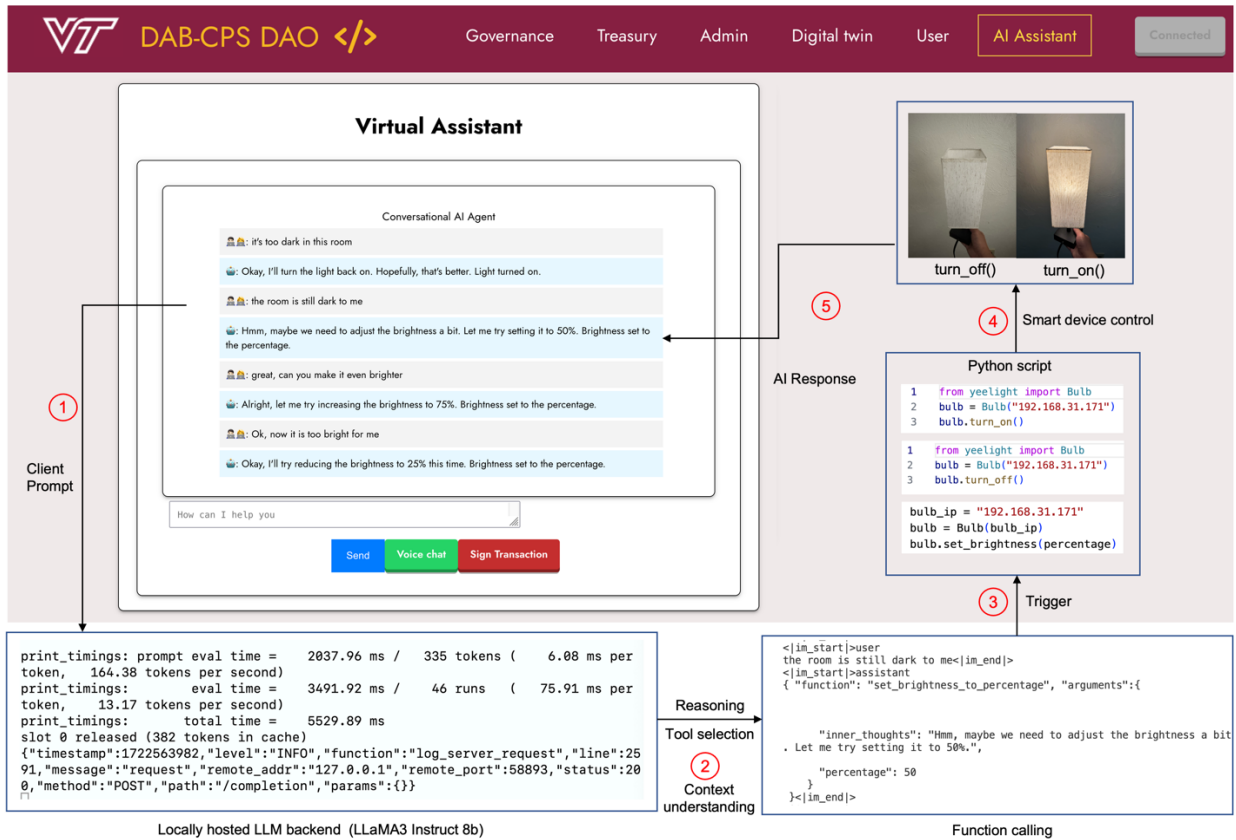

Fig. 20. Workflow of the smart appliance control using LLM-based AI virtual assistant.

### 5.4.7. Scenario 6: Autonomous building operation with LLM-based AI agent

This section aims to validate the role of the AI agent in managing smart devices for autonomous building operations through two distinct scenarios: automatic appliance control based on occupancy and autonomous adjustments according to environmental data. In the first case, we examine how the AI system adjusts appliance settings in response to occupancy changes. The detection of occupancy is a well-researched area that has been conducted through different methods, including surveillance cameras, machine learning, and sensor technologies. Therefore, the occupancy detection itself is not within the scope of this study. This experiment uses the simulated occupancy data (randomly selected between 1 and 10) and is updated every 10 minutes to test the AI agent's ability to autonomously control devices such as a smart fan and air purifier. For demonstration, this study defines low occupancy as fewer than five individuals, while high occupancy is defined as more than five individuals. As demonstrated in Fig. 21, the AI agent continuously monitors the fan and air purifier settings and occupancy data through RESTful API. Depending on the occupancy levels, the AI agent modifies the performance of the devices accordingly. If no occupants are detected, the AI agent will turn off all devices. In the case of low occupancy, it reduces the performance to a lower setting, while in high occupancy scenarios, it increases the devices' performance to meet the demand. Initially, both smart fan and air purifier were set to low performance at level 1. During the experiment, the simulated occupancy was ten individuals, which led the AI agent to perform the contextual reasoning and raise the performance settings of the fan and air purifier to level 3 and level 7, respectively, to ensure environmental comfort in the physical space.

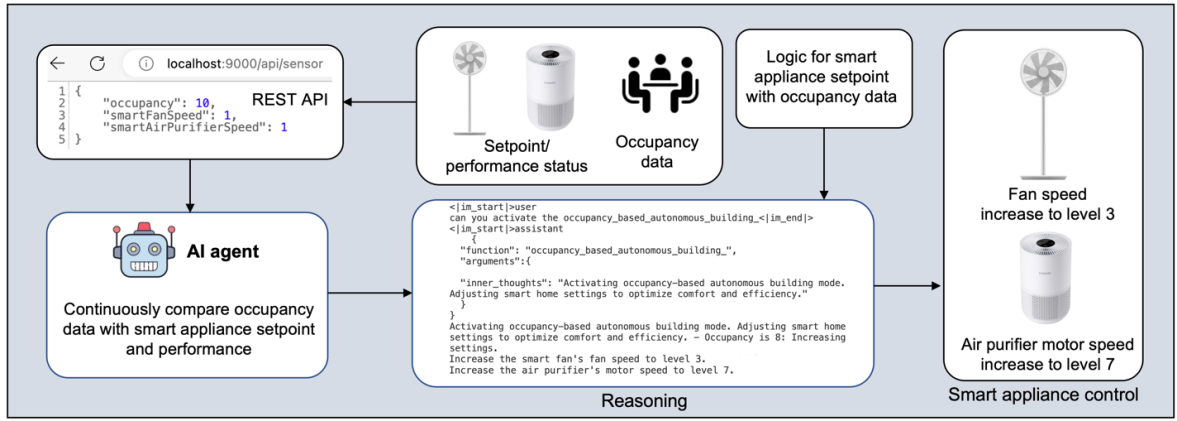

Fig. 21. Occupancy-based autonomous smart appliance control using LLM-based AI agent.

The second case aims to validate the AI agent's ability to autonomously adjust smart appliances based on baseline environmental comfort parameters retrieved from blockchain smart contracts and real-time environmental conditions, such as temperature, humidity, luminance, and carbon monoxide levels. In this scenario, the smart appliance setpoints were initially configured at their lowest levels. As shown in Fig. 22, the AI retrieved the baseline comfort parameters via web3.py, which included temperature (20°C min, 27°C max), illuminance (50-150 lux), humidity (40-100%), and carbon monoxide concentration (400-1000 ppm). The real-time room conditions—28°C temperature, 45% humidity, 34 lux, and 752 ppm carbon monoxide level—were obtained through REST API. Upon processing this data, the AI agent determined that the temperature and luminance were outside the comfort range and automatically adjusted the smart fan to speed level 3 and the smart light to 90% brightness while leaving other appliance settings unchanged. This experiment demonstrates the AI's ability to autonomously regulate the indoor environment by adjusting smart devices based on real-time data and predefined comfort thresholds. In addition, the smart appliance control by user preference is demonstrated in the previous section.

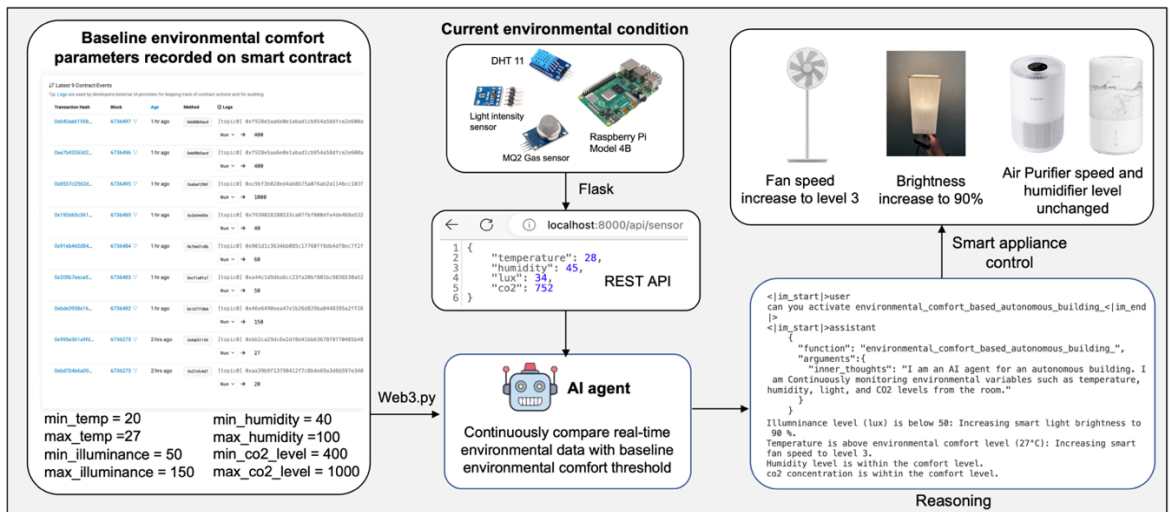

Fig. 22. Autonomous smart appliance control using environmental comfort threshold and LLM-based AI agent.

## 6. Results, evaluation, and discussion

The DAB-CPS prototype was subjected to a two-week experimental testing phase. By the end of the period, a total of 9 space bookings were successfully made through the decentralized space reservation system, generating a total revenue of 0.09 ETH for DA BCPS's DAO. On the decentralized governance platform, eight proposals are being submitted for voting. These proposals included actions such as adding and removing members, paying utility bills, sending governance tokens, and adjusting environmental parameter thresholds for smart space management. Of these, five proposals were successfully passed, while three proposals failed due to insufficient support from the DAO members, where governance token backing did not surpass the 50% threshold. This demonstrates that the voting mechanism functioned as intended by enforcing democratic decision-making in the decentralized governance process. During this period, the smart plug energy meters documented a total electricity consumption of 22.73 kilowatt-hours (kWh) for the smart appliance usage. Given the current electricity rate of $0.169475 per kWh, the calculated total electricity cost amounts to USD 3.85, which is equivalent to 0.0016 ETH. Based on this data, DAO members have initiated a proposal and subsequent voting process to authorize the transfer of 0.0016 ETH for the electricity bill payment. Additionally, throughout the period, the AI virtual assistant processed a total of 35 requests, which included 21 requests for smart home device control and 14 requests to execute blockchain-related tasks. The AI agent is also capable of autonomously adjusting smart appliances based on given comfort threshold and occupancy data to maintain optimal comfort. Table 2 presents the criteria used to evaluate the workability of the DAB-CPS system prototypes, all of which were successfully achieved. The system demonstrated its capability across multiple key areas, including user reservation of physical spaces, digital twin visualization environmental conditions and reservation status, and the effective execution of decentralized governance processes through DAO member proposals and voting. Additionally, the AI virtual assistant proved successful in both smart building control and blockchain-related task execution.

Table 2. Workability test result

| No. | Criteria | Result |
|---|---|---|
| 1 | User interaction with the physical space reservation portal | Achieved |
| 2 | Digital building twin for environmental condition and reservation status | Achieved |
| 3 | DAO members interaction with the decentralized governance platform (proposing/voting/queuing and executing the proposal) | Achieved |
| 4 | AI virtual assistant for smart building system control | Achieved |
| 5 | AI virtual assistant for blockchain related task execution | Achieved |
| 6 | AI agent for autonomous building operation | Achieved |

### 6.1. Cost analysis

Transaction on the Ethereum network requires a gas fee to compensate for the computational power needed to process the transaction on the network. The cost analysis of the DAB-CPS system evaluates the gas consumption and transaction fees for various blockchain operations performed during the experiment, including smart contract deployments, governance actions, space reservations, and AI-assistant-related tasks. These fees are expressed in Ether (ETH) and converted to USD to illustrate the financial implications of each transaction. Gas fees are measured in units of gas, with the total transaction cost calculated by multiplying the gas used by the gas price. The key operations listed in Table 3 incurred different transaction fees. For example, the deployment of core smart contracts like Governor, Timelock, and Tokens consumed significant gas, costing around 0.051903 ETH in total (~122.38 USD). Other operations, such as DAO member registration, space reservations, governance/Ethereum tokens transfer, as well as governance proposals submission, voting, queuing, and execution, had costs ranging from 0.000110 ETH (~0.26 USD) to 0.001052ETH (~2.47 USD) per transaction. It's crucial to recognize that these costs depend on the particular blockchain network selected for Dapp implementation, and transaction costs may differ when applied to a Mainnet network. The fees were calculated during testing on the Sepolia testnet and are summarized in the final column of Table 3.

Table 3. The transaction cost of the proposed decentralized governance platform.

| Operations | Smart contract | Gas | Transaction fee (ETH) | Transaction fee (USD) |
| --- | --- | --- | --- | --- |
| Contract deployment | DAO Governor | 3,880,388 | 0.003880 | 9.15 |
| Contract deployment | Timelock controller | 1,909,795 | 0.001909 | 4.50 |
| Contract deployment | GovernanceToken | 1,971,098 | 0.001971 | 4.65 |
| Contract deployment | Facilities automation | 488,638 | 0.011985 | 28.26 |
| Contract deployment | Space reservation | 1,662,788 | 0.032158 | 75.82 |
| Adding DAO member | DAO Governor | 73,610 | 0.000110 | 0.26 |
| Space Reservation payment | Space reservation | 181,123 | 0.003839 | 9.05 |
| Proposal submission | DAO Governor | 108,168 | 0.000199 | 0.47 |
| Voting on proposal | DAO Governor | 93,186 | 0.000169 | 0.40 |
| Queuing proposal | DAO Governor | 123,769 | 0.000235 | 0.38 |
| Executing the Proposal | DAO Governor | 132,563 | 0.000238 | 0.56 |
| Governance Tokens transfer | GovernanceToken | 72,954 | 0.000139 | 0.3286 |
| Ethereum tokens transfer | Timelock controller | 21,055 | 0.001052 | 2.479 |

### 6.2. Scalability

The scalability of the developed DAB-CPS prototype is evaluated by assessing both its underlying blockchain infrastructure and the LLM-based AI system. On the blockchain side, the system scalability is influenced by the limitations of the Ethereum blockchain, specifically its proof-of-stake consensus mechanism, which can limit the transaction throughput. Every transaction within the Ethereum network must receive validation from all participating nodes before being added to the blockchain. As transaction volumes rise, more nodes will be necessary to ensure network efficiency. However, with the expansion of the network, the processes required to reach a consensus also escalate, potentially resulting in delays and increased gas fees [60]. This challenge is common within the Ethereum blockchain-based systems, where throughput is limited to around 30 transactions per second [61]. However, in the proposed DAB-CPS, users interact with the decentralized governance platform by proposing ideas, voting on them, and executing approved proposals where each of these actions triggers specific smart contract functions. While this can introduce a significant number of transactions, the decentralized nature of governance helps distribute activity over time. It is quite improbable that all DAO members will simultaneously submit proposals, vote, or execute actions, which in turn reduce the likelihood of bottlenecks. Likewise, it is less likely that all users will attempt to book rooms simultaneously. These factors help mitigate potential scalability concerns in our experimental setup, which currently can process a manageable number of transactions without issue. However, if the system were to be adopted in a real-world scenario with a larger number of users, migrating to a more scalable blockchain solution like Polygon, which can process over 65,000 transactions per second [62], could be a practical solution.

Furthermore, the scalability of the proposed LLM-based AI system is evaluated based on its throughput and ability to handle concurrent user requests, specifically measuring how many requests the AI can process simultaneously and how quickly the system can respond to user queries. In our experiment, we used LlamaBench, an open-source tool for benchmarking LLM, to assess the performance of the proposed AI-based agent and virtual assistant. The results indicated that execution time and throughput varied based on the specific task. For chat or text generation, the average throughput was 33.66 tokens per second. One token is approximately equivalent to 4 English characters, and 1,500 words correspond to around 2048 tokens [63]. Smart home control tasks require longer processing time, with an average execution time of 5402.62 milliseconds per task and a throughput of 12.77 tokens per second. Blockchain-related tasks had the highest execution time, averaging 8714.91 milliseconds per task, with a throughput of 12.52 tokens per second. Text generation typically requires less computational effort compared to smart home or blockchain tasks, where the model must perform more intricate operations, resulting in longer execution times. However, the overall performance remains within acceptable limits, with even the most complex blockchain tasks completed in under 9 seconds and simpler tasks executed in as little as 5 seconds.

Concurrency user request is also an important indicator of the LLM model's scalability [64]. In this study, we used Llamacpp for model deployment, which allows parallelization based on the model context length. For instance, a LLAMA 3 8B model with a context window is 4096 tokens deployed on a machine with one 48G L40/L40s GPU that can handle up to 16 concurrent requests [65]. Although the Llama 3 model we used supports a context length of up to 128k tokens, we limited it to 4096 tokens due to limited computational resources. For real-world deployment, especially with a larger user base, we can improve the scalability by opting for models with larger context lengths and running them on machines with greater GPU RAM capacity.

### 6.3. Data Security, Privacy, and Integrity

One of the primary concerns in permissionless blockchain systems is preserving users' confidentiality. The DAB-CPS system addresses this by leveraging Ethereum's pseudonymous structure, where user identities are protected through public-key cryptography [66]. This ensures that all actions, including voting, submitting proposals, and reserving spaces, are linked to pseudonymous public keys rather than personal information. Although transactions are publicly visible on the blockchain, the identities of participants remain anonymous and secure. In addition, all participants, including DAO members, building occupants, or system users, sign transactions using private keys. This guarantees that only authorized individuals can validate and execute the transactions. The combination of public-key cryptography and digital signatures enhances security, making stored records immutable and tamper-proof once recorded on the blockchain. This strengthens user privacy, ensuring that personal data remains protected.

Another important aspect of the system's security is the accessibility of transaction data. In the DAB-CPS system, governance proposals and user votes are intentionally made public to promote transparency. The number of governance tokens held by each DAO member is also publicly accessible, fostering accountability and trust within the community. This transparency encourages active participation, as members can verify actions and engage in governance based on reliable information. Additionally, the system ensures the availability of this data permanently, thereby maintaining data integrity and preserving trust throughout the governance process.

### 6.4. Methodology evaluation

Hevner et al. [41] present seven guidelines for evaluating methodological rigor and relevance in Design Science Research (DSR). The findings in Table 4 illustrate that the development process of the DAB-CPS framework and prototype is scientific.

Table 4. Evaluation of the methodology in accordance with DSR guidelines.

| Guideline | Description |
| --- | --- |
| Design as an Artefact | This paper proposed the decentralized autonomous building cyber-physical system (DAB-CPS) framework and develop the corresponding prototype using different technical component including DAO and LLM-based AI agent and virtual assistant |
| Problem Relevance | The research addresses relevant problems and knowledge gaps identified through a literature review, including limited studies on autonomous building infrastructure, lack of research on decentralized governance of building operation, and the lack of open-source LLM application as well as DAO and AI integration in the smart building domain. |
| Design Evaluation | The methodology includes both quantitative and qualitative evaluations of the DAB-CPS framework including analysis on the system cost and scalability, methodology evaluation, as well as data security, privacy, and integrity. The system workability is also validated with six different scenarios. The results indicate that the prototype's performance meets acceptable standards. |
| Research Contributions | The research contributions can be summarized as follows:<br>• Novel DAO and AI-based decentralized governance model for smart, operational, and financially autonomous infrastructure<br>• Full-stack, open-source Dapp template for decentralized governance.<br>• Integration of DAO and LLM-based AI system and digital building twin<br>• Implementation and evaluation of the DAB-CPS prototype in the real-world settings. |
| Research Rigor | The research follows a structured DSR methodology with six clearly defined stages, from problem identification to communication of results. The prototype of the proposed DAB-CPS framework is validated and evaluated in real physical building to demonstrate its feasibility |
| Design as a Search Process | This study explored existing literature and industry practices to identify knowledge gaps in autonomous building research. The innovative DAB-CPS framework and prototype was develop using state of the art AI, digital twin and blockchain technologies. |
| Communication of Research | The implementation code base DAB-CPS prototype is publicly shared. Additionally, the research findings, prototypes design, development methodology, and evaluation results will be published in international academic journals. |

### 6.5. Limitations

The DAB-CPS framework presents a novel approach to future smart and autonomous building infrastructures. However, despite its advantages, the framework is not without limitations. This section outlines key constraints related to the proposed system and its proof-of-concept implementation.

A primary constraint is the inherent volatility of the Ethereum cryptocurrency used for transactions within the system. This volatility introduces financial instability, complicating both the physical reservations and utility/expense payments for users and DAO members. This can lead to discrepancies between expected and actual costs, potentially deterring users from widespread adoption. To address this issue, future iterations of the framework could explore the integration of stablecoins, such as USDT or USDC [67], which offer more stable decentralized payment options by being pegged to reserve assets like the U.S. Dollar. Additionally, the reliance on smart appliances and smart plugs for energy meters instead of a fully integrated smart building automation system represents a limitation of the current implementation. Despite this constraint, these devices were suitable for the research objectives and effectively demonstrated the capabilities of the DAB-CPS framework in simulating a range of building operation management functionalities. Another significant limitation lies in the technical constraints preventing AI virtual assistants from directly executing blockchain-related tasks. For security reasons, the current system avoids the direct embedding of private keys onto the AI virtual assistant and necessitates manual signing and confirmation of transactions for each user request, which, to some extent, reduces the overall efficiency and increases system complexity. Future research should focus on developing a personalized AI system capable of automated blockchain task execution without security concerns. This could potentially involve the integration of zero-knowledge proofs [68] with the LLM, which could allow users to prove their identity to the AI system without revealing their private keys, thereby maintaining security while enabling more automated interactions.

### 6.6. Future outlooks

As per the authors' knowledge, this study represents the initial attempt to develop an integrative framework and prototype for autonomous building infrastructure that synergizes artificial intelligence, digital twin, and distributed ledger technology. The author believes that the DAB-CPS framework and its associated prototypes not only have the potential to significantly advance the current state of research and knowledge in autonomous buildings but could also contribute to future research on automation in smart communities and cities.

The framework's potential applications are not limited to building infrastructure. The decentralized governance platform could also be customized to govern civil infrastructure, where entire systems could be democratically controlled by DAO members and autonomously operated. This could revolutionize how we manage and interact with urban environments, from transportation networks to utility systems, fostering a new era of citizen participation and efficient resource management. In addition, LLMs play a crucial role in the proposed AI system, which is the backbone of this study. The current trend towards smaller and more efficient language models, as evidenced by Microsoft's Phi-3 [69], Meta's LLaMA 3.2 model [70] , and Google's Gemma-2 model [71], indicates that powerful AI systems will be able to effectively deploy on low-cost edge devices such as Raspberry Pi while offering impressive performance. This development could significantly advance research in smart cities and infrastructure by having advanced AI systems to be more accessible and have widespread deployment across urban environments. Furthermore, LLMs with multi-modal capabilities present an exciting opportunity for enhancing smart building functionalities. These advanced AI systems are capable of processing visual, auditory, and textual data, which could revolutionize tasks such as surveillance, emergency response, and safety detection. The contextual awareness of these models could significantly advance research in autonomous buildings, leading to more sophisticated and responsive urban environments.

### 7. Conclusion

This paper presents a novel Decentralized Autonomous Building Cyber-Physical System (DAB-CPS), an innovative, integrative framework for smart buildings, which is comprised of web3-based governance, artificial intelligence, digital building twin, facilities management, and building automation systems. The framework aims to serve as a blueprint for a self-governing, autonomous building infrastructure by leveraging blockchain technology, DAOs, and LLMs-powered building automation systems. The DAB-CPS framework is designed to create a financially self-sustaining building infrastructure capable of autonomously managing its operations, including generating revenue for its operational expense, allowing the building to self-sustain and self-managed in a decentralized manner.

The DAB-CPS framework comprises several key components. The decentralized governance platform, powered by DAO's governance, facilitates transparent decision-making and resource management. The space reservation system allows users to reserve physical space autonomously using blockchain technology. The digital twin component provides real-time visualization of both the reservation statuses and environmental conditions such as temperature, humidity, and occupancy. The large language-powered AI systems allow users to interact with the building through voice and text interfaces for blockchain and facility management-related tasks such as smart appliance control and blockchain transactions. The AI agent also powers the autonomous building operations by autonomously adjusting smart appliances such as lighting and HVAC based on occupancy and the baseline environmental comfort threshold to maintain optimal conditions for occupants. The resource and code implementation for these components is available on a GitHub repository under an open-source license, allowing for further development and application of this framework beyond autonomous building management.

In this study, the prototype of the DAB-CPS framework was conducted in a real-world building to validate its practical application. Evaluations of the system included analyses of cost efficiency, scalability of the AI and governance system, as well as data security, privacy, and system integrity. The workability of the DAB-CPS system was validated through six different scenarios, including user interactions with the space reservation portal, income and expense management handled by the DAO entity, proposal voting conducted by DAO members for building-related decisions, and the AI virtual assistant's performance in carrying out blockchain-related tasks and facility management operations, such as controlling smart appliances based on environmental data. The results from these evaluations demonstrated that the developed prototype system completed these operations and can potentially serve as the viable framework for autonomous building operation and management in building infrastructure.

**CRediT authorship contribution statement**
**Reachsak Ly**: Writing – review & editing, Writing – original draft, Visualization, Conceptualization.
**Alireza Shojaei**: Project administration, Supervision, Conceptualization, Methodology.

**Declaration of Generative AI and AI-assisted technologies in the writing process.**
During the preparation of this work, the author(s) used OpenAI GPT4 to improve readability and language. After using this tool/service, the author(s) reviewed and edited the content as needed and take(s) full responsibility for the content of the publication.

**Declaration of Competing Interest**
The authors declare that they have no known competing financial interests or personal relationships that could have appeared to influence the research presented in this paper.

**Data availability**
The code is publicly available in a GitHub repository under an open-source license.